\documentclass[10pt,twocolumn,letterpaper]{article}

\usepackage{cvpr}
\usepackage{times}
\usepackage{epsfig}
\usepackage{graphicx}
\usepackage{amsmath}
\usepackage{amssymb}
\usepackage{tabularx}

\usepackage{bm}
\usepackage{booktabs}

\newcommand{\STAB}[1]{\begin{tabular}{@{}c@{}}#1\end{tabular}}
\newcommand{\STABBa}[1]{\STAB{\rotatebox[origin=l]{90}{\parbox[c]{1.2cm}{#1}}}}

\usepackage[pagebackref=true,breaklinks=true,letterpaper=true,colorlinks,bookmarks=false]{hyperref}

\cvprfinalcopy % *** Uncomment this line for the final submission

\def\cvprPaperID{4999} % *** Enter the CVPR Paper ID here

\ifcvprfinal\fi

\begin{document}

%-------------------------------------------------------------------------
% Title
\title{Learning View Priors for Single-view 3D Reconstruction}

\author{
    Hiroharu Kato${}^\text{1}$ and Tatsuya Harada${}^\text{1,2}$\\
    ${}^\text{1}$The University of Tokyo, ${}^\text{2}$RIKEN\\
    {\tt\small \{kato,harada\}@mi.t.u-tokyo.ac.jp}
}

\maketitle
\thispagestyle{empty}

%-------------------------------------------------------------------------
% Abstract
\begin{abstract}
There is some ambiguity in the 3D shape of an object when the number of observed views is small. Because of this ambiguity, although a 3D object reconstructor can be trained using a single view or a few views per object, reconstructed shapes only fit the observed views and appear incorrect from the unobserved viewpoints. To reconstruct shapes that look reasonable from any viewpoint, we propose to train a discriminator that learns prior knowledge regarding possible views. The discriminator is trained to distinguish the reconstructed views of the observed viewpoints from those of the unobserved viewpoints. The reconstructor is trained to correct unobserved views by fooling the discriminator. Our method outperforms current state-of-the-art methods on both synthetic and natural image datasets; this validates the effectiveness of our method.

\end{abstract}

%-------------------------------------------------------------------------

% Introduction
\section{Introduction}
\label{sec:introduction}
\vspace{-1mm}

Humans can estimate the 3D structure of an object in a single glance. We utilize this ability to grasp objects, avoid obstacles, create 3D models using CAD, and so on. This is possible because we have gained prior knowledge about the shapes of 3D objects.
% For example, humans can estimate the 3D structure of a chair in the left image in Figure~\ref{fig:top} as the shape shown in the lower half of the figure.

Can machines also acquire this ability? This problem is called {\it single-view 3D object reconstruction} in computer vision. A straightforward approach is to train a reconstructor using 2D images and their corresponding ground truth 3D models~\cite{choy20163d,fan2016point,gwak2017weakly,Jiang_2018_ECCV,kar2017learning,richter2018matryoshka,tatarchenko2017octree}. However, creating 3D annotations requires extraordinary effort from professional 3D designers. Another approach is to train a reconstructor using a single view or multiple views of an object without explicit 3D supervision~\cite{kanazawa2018learning,kar2015category,kato2018neural,tulsiani2017multi,yan2016perspective}. We call this approach {\it view-based training}. This approach typically requires annotations of silhouettes of objects and viewpoints, which are relatively easy to obtain.

\begin{figure}[t]
    \begin{center}
    \includegraphics[width=1.0\linewidth,bb=0 0 396 283]{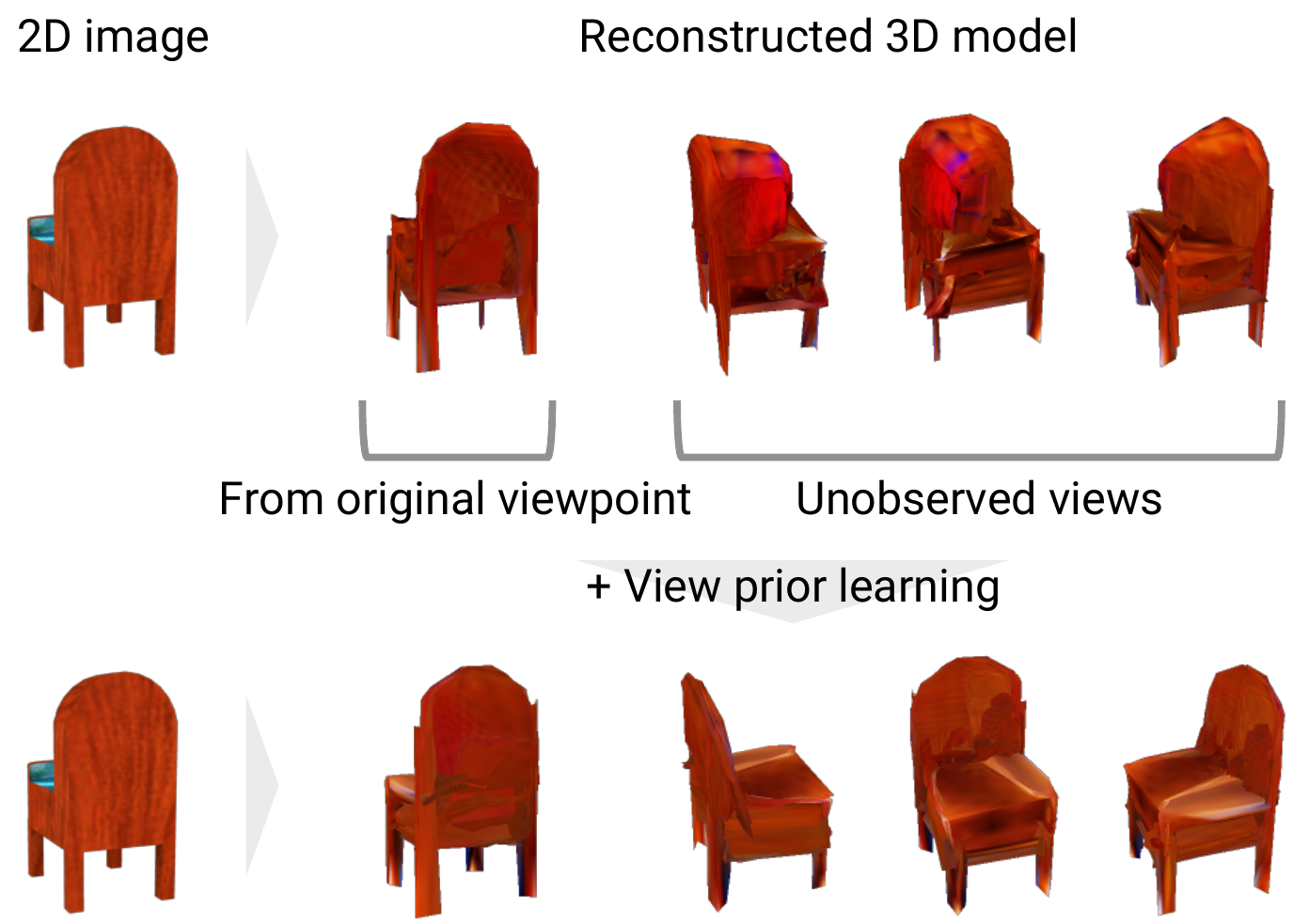}
    \end{center}
       \caption{When a 3D reconstructor is trained using only a single view per object, because of ambiguity in the 3D shape of an object, it reconstructs a shape which only fits the observed view and looks incorrect from unobserved viewpoints (upper). By introducing a discriminator that learns prior knowledge of correct views, the reconstructor is able to generate a shape that is viewed as reasonable from any viewpoint (lower). }
    \label{fig:top}
    \vspace{-2mm}
\end{figure}

\begin{table*}[t]
    \begin{center}
        \small
        \begin{tabular}{llll}
            \toprule
                & Work                                 & Class A                                               & Class B \\
            \hline
            (a) & \cite{Jiang_2018_ECCV,yang2018dense} & Predicted 3D shapes                                   & Their corresponding ground truth 3D shapes \\
            (b) & \cite{gwak2017weakly,wu2018learning} & Predicted 3D shapes                                   & 3D shape collections \\
            (c) & Ours                                 & Views of predicted 3D shapes from observed viewpoints & Views of predicted 3D shapes from random viewpoints \\
            (d) & -                                    & Views of predicted 3D shapes                          & Views in a training dataset \\
            \bottomrule
        \end{tabular}
    \end{center}
    \caption{Summary of discriminators in learning-based 3D reconstruction. Discriminator (d) is described in Section~\ref{sec:method_another_discriminator}.}
    \label{table:discriminators}
    \vspace{-2mm}
\end{table*}

Because a ground truth 3D shape is not given in view-based training, there is some ambiguity in the possible shapes. In other words, several different 3D shapes can be projected into the same 2D view, as shown in the upper half and lower half of Figure~\ref{fig:top}. To reduce this ambiguity, twenty or more views per object are typically used in training ~\cite{kato2018neural,yan2016perspective}. However, this is not practical in many cases in terms of feasibility and scalability. When creating a dataset by taking photos, if an object is moving or deforming, it is difficult to take photos from many viewpoints. In addition, when creating a dataset using a large number of photos from the Internet, it is not always possible to collect multiple views of an object. Therefore, it is desirable that a reconstructor can be trained using a few views or even a single view of an object.

In this work, we focus on training a reconstructor using a single view or a few views for single-view 3D object reconstruction. In this case, the ambiguity in shapes in training is not negligible. The upper half of Figure~\ref{fig:top} shows the result of single-view 3D reconstruction using a conventional approach~\cite{kato2018neural}. Although this method originally uses multiple views per object for training, a single view is used in this experiment. As a result, the reconstructed shape looks correct when viewed from the same viewpoint as the input image, however, it looks incorrect from other viewpoints. This is because the reconstructor is unaware of unobserved views and generates shapes that only fit the observed views.

How can a reconstructor overcome shape ambiguity and correctly estimate shapes? The hint is shown in Figure~\ref{fig:top}. Humans can recognize that the three views of the chair in the upper-right of Figure~\ref{fig:top} are incorrect because we have prior knowledge of how a chair looks, having seen many chairs in the past. If machines also have knowledge regarding the correct views, they would use it to estimate shapes more accurately.

We implement this idea on machines by using a discriminator and adversarial training~\cite{goodfellow2014generative}. One can see from the upper half of Figure~\ref{fig:top} that, with the conventional method, views of estimated shapes from observed viewpoints converge to the correct views, while unobserved views do not always become correct. Therefore, we train the discriminator to distinguish the observed views of estimated shapes from the unobserved views. This results in the discriminator obtaining knowledge regarding the correct views. By training the reconstructor to fool the discriminator, reconstructed shapes from all viewpoints become indistinguishable and to be viewed as reasonable from any viewpoint. The lower half of Figure~\ref{fig:top} shows the results from the proposed method.

Learning prior knowledge of 3D shapes using 3D models was tackled in other publications~\cite{gwak2017weakly,wu2018learning}. In contrast, we focus on prior knowledge of 2D views rather than 3D shapes. Because our method does not require any 3D models for training, ours can scale to various categories where 3D models are difficult to obtain.

The major contributions can be summarized as follows.
\begin{itemize}
    \setlength\itemsep{0em}
    \item In view-based training of single-view 3D reconstruction, we propose a method to predict shapes which are viewed as reasonable from any viewpoint by learning prior knowledge of object views using a discriminator. Our method does not require any 3D models for training.
    \item We conducted experiments on both synthetic and natural image datasets and we observed a significant performance improvement for both datasets. The advantages and limitations of the method are also examined via extensive experimentation.
\end{itemize}

\section{Related work}
\vspace{-1mm}

A simple and popular approach for learning-based 3D reconstruction is to use 3D annotations. Recent studies focus on integrating multiple views~\cite{choy20163d,kar2017learning}, memory efficiency problem of voxels~\cite{tatarchenko2017octree}, point cloud generation~\cite{fan2016point}, mesh generation~\cite{groueix2018atlasnet,wang2018pixel2mesh}, advanced loss functions~\cite{Jiang_2018_ECCV}, and neural network architectures~\cite{richter2018matryoshka}.

To reduce the cost of 3D annotation, view-based training has recently become an active research area. The key of training is to define a differentiable loss function for view reconstruction. A loss function of silhouettes using chamfer distance~\cite{kar2015category}, differentiable projection of voxels~\cite{tulsiani2017multi,wu2017marrnet,yan2016perspective,zhu2017rethinking}, point clouds~\cite{insafutdinov2018unsupervised,lin2017learning}, and meshes~\cite{kato2018neural} is proposed. Instead of using view reconstruction, 3D shapes can be reconstructed via view synthesis~\cite{tatarchenko2016multi}.

As mentioned in the previous section, it is not easy to train reconstructors using a small number of views. For this problem, some methods use human knowledge of shapes as regularizers or constraints. For example, the graph Laplacian of meshes was regularized~\cite{kanazawa2018learning,wang2018pixel2mesh}, and shapes were assumed to be symmetric~\cite{kanazawa2018learning}. Instead of using manually-designed constraints, others attempted to acquire prior knowledge of shapes from data. Learning category-specific mean shapes~\cite{kanazawa2018learning,kar2015category} is an example. Adversarial training is another way to learn shape priors. Yang \etal~\cite{yang2018dense} and Jiang \etal~\cite{Jiang_2018_ECCV} used discriminators on an estimated shape and its corresponding ground truth shape to make the estimated shapes more realistic. Gwak \etal~\cite{gwak2017weakly} and Wu \etal~\cite{wu2018learning} used discriminators on generated shapes and a shape collection. In contrast, our method does not require 3D models to learn prior knowledge. Table~\ref{table:discriminators} lists a summary of these discriminators.

%-------------------------------------------------------------------------
\section{View-based training of single-view 3D object reconstructors with view prior learning}
\vspace{-1mm}

\begin{figure*}[t]
    \begin{center}
    \includegraphics[width=500px,bb=0 0 681 167]{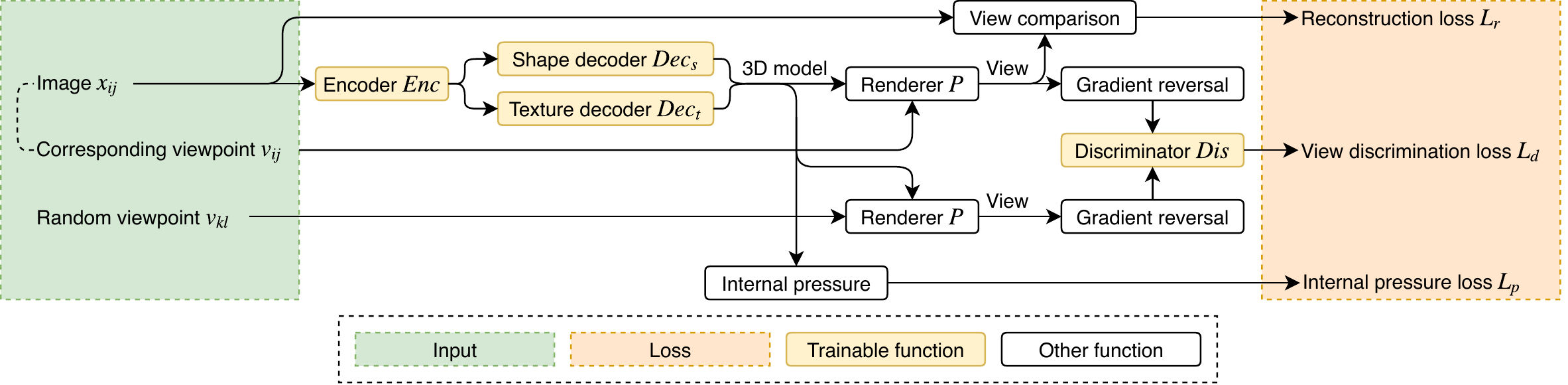}
    \end{center}
       \caption{Architecture of the proposed method. The main point of our method is the use of discrimination loss to learn priors of views. While the discriminator aims to minimize discrimination loss, the encoder and decoders try to maximize it using a gradient reversal layer.}
    \label{fig:method1}
    % \vspace{-3mm}
\end{figure*}

\begin{figure}[t]
    \begin{center}
    \includegraphics[width=228px,bb=0 0 310 88]{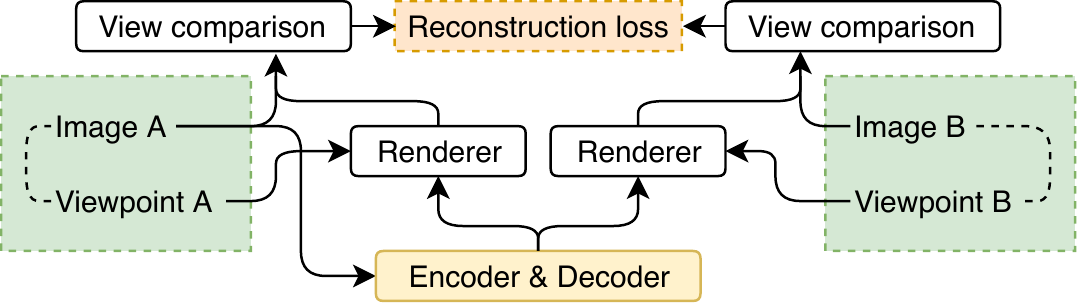}
    \end{center}
       \caption{Reconstruction loss in multi-view training. Images A and B are views of the same object. Although only the loss with respect to a view reconstructed from image A is shown in this figure, the loss with respect to image B is also computed.}
    \label{fig:method2}
    % \vspace{-4mm}
\end{figure}

In this section, we introduce a simple view-based method to train 3D reconstructors based on~\cite{kato2018neural}. Then, we describe our main technique, called {\it view prior learning (VPL)}. We also explain a technique to further improve reconstruction accuracy by applying internal pressure to shapes. Figure~\ref{fig:method1} shows the architecture of our method.

For training, our method requires a dataset that contains single or multiple views of objects, and their silhouette and viewpoint annotations, similar to previous studies~\cite{kanazawa2018learning,kato2018neural,tulsiani2017multi,yan2016perspective}. Additionally, ours can also use class labels of views if they are available. After training, reconstruction is performed without silhouette, viewpoint, and class label annotations.

%-------------------------------------------------------------------------
\subsection{View-based training for 3D reconstruction}
\label{sec:view_based_training}
\vspace{-1mm}

In this section, we describe our baseline method for 3D reconstruction. We extend a method which uses silhouettes in training~\cite{kato2018neural} to handle textures using a texture decoder and perceptual loss~\cite{johnson2016perceptual}.

\vspace{-3mm}
\paragraph{Overview.}
The common approach to view-based training of 3D reconstructors is to minimize the difference between views of a reconstructed shape and views of a ground truth shape. Let $x_{ij}$ be the view of an object $o_i$ from a viewpoint $v_{ij}$, $N_o$ be the number of objects in the training dataset, $N_v$ be the number of viewpoints per object, $R(\cdot)$ be a reconstructor that takes an image and outputs a 3D model, $P(\cdot, \cdot)$ be a renderer that takes a 3D model and a viewpoint and outputs the view of the given model from the given viewpoint, and $\mathcal{L}_v(\cdot, \cdot)$ be a function that measures the difference between two views. Then, reconstruction loss is defined as
\begin{eqnarray}
  \mathcal{L}_r (x, v) = \sum_{i = 1}^{N_o} \sum_{j = 1}^{N_v} \sum_{k = 1}^{N_v} \mathcal{L}_v(P(R(x_{ij}), v_{ik}), x_{ik}).
\end{eqnarray}
We call the case where $N_v = 1$ {\it single-view training}. In this case, the reconstruction loss is simplified to $\mathcal{L}_r (x, v) = \sum_{i = 1}^{N_o} \mathcal{L}_v(P(R(x_{i1}), v_{i1}), x_{i1})$. We call the case where $2 \leq N_v$ {\it multi-view training}.

\vspace{-3mm}
\paragraph{3D representation and renderer.}
Some works use voxels as a 3D representation in view-based training~\cite{tulsiani2017multi,yan2016perspective}. However, voxels are not well suited to view-based training because using high-resolution views of voxels is difficult as voxels are memory inefficient. Recently, this problem was overcome by Kato \etal~\cite{kato2018neural} by using a mesh as a 3D representation and a differentiable mesh renderer. Following this work, we also use a mesh and their renderer\footnote{We modified the approximate differentiation of the renderer. Details are described in the supplementary material.}.

\vspace{-3mm}
\paragraph{Reconstructor.}
In this work, a 3D model is represented by a pair of a shape and a texture. Our reconstructor $R(\cdot)$ uses an encoder-decoder architecture. An encoder $\mathit{Enc}(\cdot)$ encodes an input image, and a shape decoder $\mathit{Dec}_s(\cdot)$ and texture decoder $\mathit{Dec}_t(\cdot)$ generate a 3D mesh and a texture image, respectively. Following recent learning-based mesh reconstruction methods~\cite{kanazawa2018learning,kato2018neural,wang2018pixel2mesh}, we generate a shape by moving the vertices of a pre-defined mesh. Therefore, the output of the shape decoder is the coordinates of the estimated vertices. The details of the encoder and the decoders are described in the supplementary material.

\vspace{-3mm}
\paragraph{View comparison function.}
Color images (RGB channels) and silhouettes (alpha channels) are processed separately in $\mathcal{L}_v(\cdot, \cdot)$. Let $x$ and $\hat{x}=P(R(x), v)$ be a ground truth view and an estimated view, $x_c$, $\hat{x}_c$ be the RGB channels of $x$, $\hat{x}$, and $x_s, \hat{x}_s$ be the alpha channels of $x$, $\hat{x}$. The silhouette at the $i$-th pixel $x_{s_i}$ is set to one if an object exists at the pixel and to zero if the pixel is part of the background. $x_s$ can take a value between zero and one owing to anti-aliasing of the renderer. To compare color images $x_c, \hat{x}_c$, we use perceptual loss $\mathcal{L}_p$~\cite{johnson2016perceptual} with additional feature normalization. Let $F_m(\cdot)$ be the $m$-th feature map of $N_f$ maps in a pre-trained CNN for image classification. In addition, let $C_m$, $H_m$, $W_m$ be the channel size, height, and width of $F_m(\cdot)$, respectively. Specifically, we use the five feature maps after convolution layers of AlexNet~\cite{krizhevsky2012imagenet} for $F_m(\cdot)$. Then, using $D_m = C_m H_m W_m$, the perceptual loss is defined as
\begin{eqnarray}
    \mathcal{L}_c(\hat{x_c}, x_c) = \sum_{m=1}^{N_f} \frac{1}{D_m} \left| \frac{F_m(\hat{x}_c)}{| F_m(\hat{x}_c)|} - \frac{F_m(x_c)}{| F_m(x_c)|} \right|^2.
\end{eqnarray}
For silhouettes $x_s, \hat{x}_s$, we use their multi-scale cosine distance. Let $x_s^i$ be an image obtained by down-sampling $x_s$ $2^{i-1}$ times, and $N_s$ be the number of scales. We define the loss function as
\begin{eqnarray}
  \label{eq:silhouette1}
  \mathcal{L}_s(x_s, \hat{x}_s) = \sum_{i=1}^{N_s} \left( 1 - \frac{x_s^i \cdot \hat{x}_s^i}{|x_s^i| |\hat{x}_s^i|} \right).
\end{eqnarray}
We also use negative intersection over union (IoU) of silhouettes, as was used in~\cite{kato2018neural}. Let $\odot$ be an elementwise product. This loss is defined as
\begin{eqnarray}
  \label{eq:silhouette2}
  \mathcal{L}_s(x_s, \hat{x}_s) = 1 - \frac{|x_s \odot \hat{x}_s|_1}{|x_s + \hat{x}_x - x_s \odot \hat{x}_s|_1}.
\end{eqnarray}
The total reconstruction loss is $\mathcal{L}_v = \mathcal{L}_s + \lambda_c \mathcal{L}_c$. $\lambda_c$ is a hyper-parameter.

\vspace{-3mm}
\paragraph{Training.}
We optimize $R(\cdot)$ using mini-batch gradient descent. Figure~\ref{fig:method1} shows the architecture of single-view training. In multi-view training, we randomly take two views of an object in one minibatch. The architecture for computing $\mathcal{L}_r$ in this case is shown in Figure~\ref{fig:method2}.

%-------------------------------------------------------------------------
\subsection{View prior learning}
\label{sec:view_prior_learning}
\vspace{-1mm}

As described in Section~\ref{sec:introduction}, in view-based training, a reconstructor can generate a shape that looks unrealistic from unobserved viewpoints. In order to reconstruct a shape that is viewed as realistic from any viewpoint, it is necessary to (1) learn the difference between correct views and incorrect views, and (2) tell the reconstructor how to modify incorrect views. In view-based training, reconstructed views from observed viewpoints converge to the real views in a training dataset by minimizing the reconstruction loss, and views from unobserved viewpoints do not always become correct. Therefore, the former can be regarded as correct and realistic views, and the latter can be regarded as incorrect and unrealistic views. Based on this assumption, we propose to train a discriminator that distinguishes estimated views at observed viewpoints from estimated views at unobserved viewpoints to learn the correctness of views. The discriminator can pass this knowledge to the reconstructor by back-propagating the gradient of the discrimination loss into the reconstructor via estimated views and shapes as with adversarial training in image generation~\cite{goodfellow2014generative} and domain adaptation~\cite{ganin2016domain}.

Concretely, let $\mathit{Dis} (\cdot, \cdot)$ be a trainable discriminator that takes a view and its viewpoint and outputs the probability that the view is correct, and $\mathcal{V}$ be the set of all viewpoints in the training dataset. Using cross-entropy, we define {\it view disrcimination loss} as
\begin{multline}
  \mathcal{L}_d(x_{ij}, v_{ij}) = - \log (\mathit{Dis}(P(R(x_{ij}), v_{ij}), v_{ij})) \\
                                  - \sum_{v_u \in \mathcal{V}, v_u \neq v_{ij}} \frac{\log (1 - (\mathit{Dis}(P(R(x_{ij}), v_u), v_u)))}{|\mathcal{V}-1|}.
\end{multline}
In minibatch training, we sample one random view for each reconstructed object to compute $\mathcal{L}_d$.

\vspace{-3mm}
\paragraph{Stability of training.}
Although adversarial training is generally not stable, training of our proposed method is stable. It is known that training of GANs fails when the discriminator is too strong to be fooled by the generator. This problem is explained from the distinction of the supports of {\it real} and {\it fake} samples~\cite{arjovsky2017towards}. However, in our case, it is very difficult to distinguish views correctly in the earlier training stage because view reconstruction is not accurate and views are incorrect from any viewpoint. Even in the later stage, the reconstructor can easily fool the discriminator by slightly breaking the correct views. Therefore, the discriminator cannot be dominant in our method.

\vspace{-3mm}
\paragraph{Optimization of the reconstructor.}
The original procedure of adversarial training requires optimizing a discriminator and a generator iteratively~\cite{goodfellow2014generative}. Subsequently, Ganin \etal~\cite{ganin2016domain} proposed to train a generator using the reversed gradient of discrimination loss. The proposed gradient reversal layer does nothing in the forward pass, although it reverses the sign of gradients and scales them $\lambda_d$ times in the backward pass. This layer is posed on the right before a discriminator. Because this optimization procedure is not iterative, the training time is shorter than in iterative optimization. Furthermore, we experimentally found that the performance of the gradient reversal and iterative optimization are nearly the same in our problem. Therefore, we use the gradient reversal layer for training the reconstructor.

\vspace{-3mm}
\paragraph{Image type for the discriminator.}
The discriminator can take both RGBA images and silhouette images. We give it RGBA images when texture prediction is conducted, otherwise we give it silhouettes.

\vspace{-3mm}
\paragraph{Class conditioning.}
In addition, a discriminator can be conditioned on class labels using the conditional GAN~\cite{mirza2014conditional} framework. When class labels are known, view discrimination becomes easier and the discriminator becomes more reliable. We use the projection discriminator~\cite{miyato2018cgans} for class conditioning. Note that the test phase does not require class labels even in this case.

\vspace{-3mm}
\paragraph{Another possible discriminator.}
\label{sec:method_another_discriminator}
We propose to train a discriminator on views of reconstructed shapes at observed and unobserved viewpoints. Another possible approach is to distinguish reconstructed views from real views in a training dataset. In fact, this discriminator does not work well because generating a view that is difficult to distinguish from the real view is very difficult. This is caused by the limitation of the representation ability of the reconstructor and renderer. Table~\ref{table:discriminators} shows a summary of the discriminators we have explained thus far.

%-------------------------------------------------------------------------
\subsection{Internal pressure}
\vspace{-1mm}
One of the most popular methods in multi-view 3D reconstruction is visual hull~\cite{laurentini1994visual}. In visual hull, a point inside all silhouettes is assumed to be inside the object. In other words, in terms of shape ambiguity, visual hull produces the shape with the largest volume. Following this policy, we inflate the volume of the estimated shapes by giving them internal pressure in order to maximize their volume. Concretely, we add a gradient along the normal of the face for each vertex of a triangle face. Let $p_i$ be one of the vertices of a triangle face, and $n$ be the normal of the face. We add a loss term $\mathcal{L}_p$ that satisfies
$\frac{\partial \mathcal{L}_p(p_i)}{\partial p_i} = -n$.

%-------------------------------------------------------------------------
\subsection{Summary}
\vspace{-1mm}

In addition to using reconstruction loss $\mathcal{L}_r = \mathcal{L}_s + \lambda_c \mathcal{L}_c$, we propose to use view discrimination loss $\mathcal{L}_d$ to reconstruct realistic views and internal pressure loss $\mathcal{L}_p$ to inflate reconstructed shapes. The total loss is $\mathcal{L} = \mathcal{L}_s + \lambda_c \mathcal{L}_c + \mathcal{L}_d + \lambda_p \mathcal{L}_p$. The hyperparameters of loss weighting are $\lambda_c$, $\lambda_p$, and $\lambda_d$. Because $\lambda_d$ is used in the gradient reversal layer, it does not appear in $\mathcal{L}$. The entire architecture is shown in Figure~\ref{fig:method1}.

%%%%%%%%%%%%%%%%%%%%%%%%%%%%%%%%%%%%%%%%%%%%%%%%%%%%%%%%%%%%%%%%%%%%%%%%%%%%%%%%
\section{Experiments}
\vspace{-1mm}

We tested our proposed view prior learning (VPL) on synthetic and natural image datasets. We conducted an extensive evaluation of our proposed method using a synthetic dataset because it consists of a large number of objects with accurate silhouette and viewpoint annotations.

As a metric of the reconstruction accuracy, we used intersection over union (IoU) of a predicted shape and a ground truth that was used in many previous publications~\cite{choy20163d,fan2016point,kanazawa2018learning,kar2017learning,kato2018neural,richter2018matryoshka,tatarchenko2017octree,tulsiani2017multi,yan2016perspective}. To fairly compare our results with those in the literature, we computed IoU after converting a mesh into a volume of $32^3$ voxels\footnote{Another popular metric is the chamfer distance of point clouds. However, this metric is not suitable for use in view-based learning. Because it commonly assumes that points are distributed on surfaces, it is influenced by invisible structures inside shapes, which are impossible to learn in view-based training. This problem does not arise when using IoU because it commonly assumes that the interior of a shape is filled. }.

%%%%%%%%%%%%%%%%%%%%%%%%%%%%%%%%%%%%%%%%%%%%%%%%%%%%%%%%%%%%%%%%%%%%%%%%%%%%%%%%
\subsection{Synthetic dataset}
\vspace{-1mm}

As a synthetic dataset, we used ShapeNet~\cite{chang2015shapenet}, a large-scale dataset of 3D CAD models. We use $43,784$ objects in thirteen categories from ShapeNet. By using ShapeNet and a renderer, a dataset of views, silhouettes, viewpoints, and ground truth 3D shapes can be synthetically created. We used ground truth 3D shapes only for validation and testing. We used rendered views and train/val/test splits provided by Kar \etal~\cite{kar2017learning}. In this dataset, each 3D model is rendered from twenty random viewpoints. Each image has a resolution of $224 \times 224$. We augmented the training images by random color channel flipping and horizontal flipping, as was used in~\cite{kar2017learning,richter2018matryoshka}\footnote{When flipping images, we also flip the corresponding viewpoints.}. We use all or a subset of views for training, and all views were used for testing.

We used Batch Normalization~\cite{ioffe2015batch} and Spectral Normalization~\cite{miyato2018spectral} in the discriminator. The parameters were optimized with the Adam optimizer~\cite{kingma2014adam}. The architecture of the encoder and decoders, hyperparameters, and optimizers are described in the supplementary material. The hyperparameters were tuned using the validation set. We used Equation~\ref{eq:silhouette1} as the view comparison function for silhouettes.

%%%%%%%%%%%%%%%%%%%%%%%%%%%%%%%%%%%%%%%%%%%%%%%%%%%%%%%%%%%%%%%%%%%%%%%%%%%%%%%%
\subsubsection{Single-view training}
\label{sec:exp_single_view}
\vspace{-1mm}

At first, we trained reconstructors in single-view training described in Section~\ref{sec:view_based_training}. Namely, we used only one randomly selected view out of twenty views for each object in training.

\begin{figure}[t]
    \begin{center}
        \raisebox{5.5mm}{\makebox[13mm][l]{\small{Baseline}}}
        \includegraphics[trim={0.8cm 0.8cm 0.8cm 0.8cm},clip,height=13mm,width=13mm]{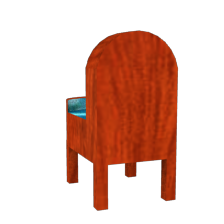}
        \includegraphics[trim={0.8cm 0.8cm 0.8cm 0.8cm},clip,height=13mm,width=13mm]{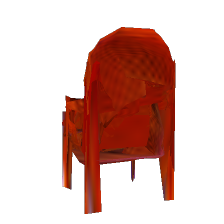}
        \includegraphics[trim={0.8cm 0.8cm 0.8cm 0.8cm},clip,height=13mm,width=13mm]{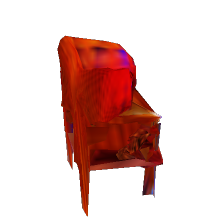}
        \includegraphics[trim={0.8cm 0.8cm 0.8cm 0.8cm},clip,height=13mm,width=13mm]{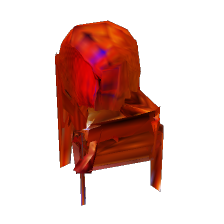}
        \includegraphics[trim={0.8cm 0.8cm 0.8cm 0.8cm},clip,height=13mm,width=13mm]{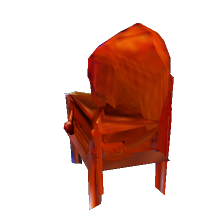} \\
        \raisebox{5.5mm}{\makebox[13mm][l]{\small{Proposed}}}
        \includegraphics[trim={0.8cm 0.8cm 0.8cm 0.8cm},clip,height=13mm,width=13mm]{./figure/images/a691eee4545ce2fade94aad0562ac2e_00.png}
        \includegraphics[trim={0.8cm 0.8cm 0.8cm 0.8cm},clip,height=13mm,width=13mm]{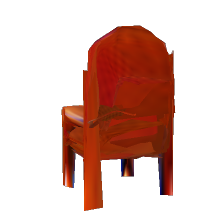}
        \includegraphics[trim={0.8cm 0.8cm 0.8cm 0.8cm},clip,height=13mm,width=13mm]{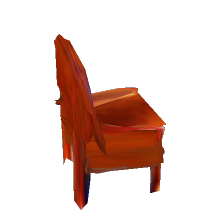}
        \includegraphics[trim={0.8cm 0.8cm 0.8cm 0.8cm},clip,height=13mm,width=13mm]{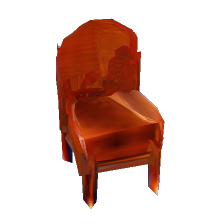}
        \includegraphics[trim={0.8cm 0.8cm 0.8cm 0.8cm},clip,height=13mm,width=13mm]{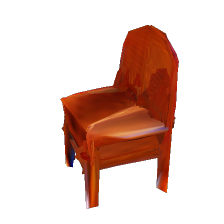} \\
        \raisebox{5.5mm}{\makebox[13mm][l]{\small{Baseline}}}
        \includegraphics[trim={0.8cm 0.8cm 0.8cm 0.8cm},clip,height=13mm,width=13mm]{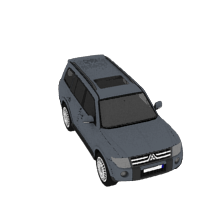}
        \includegraphics[trim={0.8cm 0.8cm 0.8cm 0.8cm},clip,height=13mm,width=13mm]{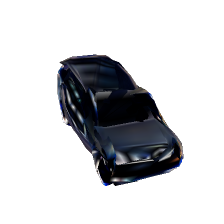}
        \includegraphics[trim={0.8cm 0.8cm 0.8cm 0.8cm},clip,height=13mm,width=13mm]{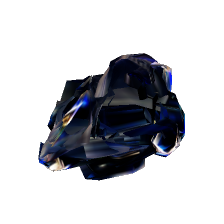}
        \includegraphics[trim={0.8cm 0.8cm 0.8cm 0.8cm},clip,height=13mm,width=13mm]{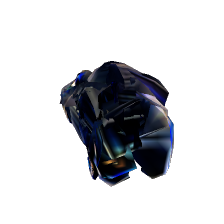}
        \includegraphics[trim={0.8cm 0.8cm 0.8cm 0.8cm},clip,height=13mm,width=13mm]{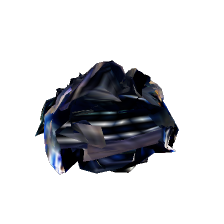} \\
        \raisebox{5.5mm}{\makebox[13mm][l]{\small{Proposed}}}
        \includegraphics[trim={0.8cm 0.8cm 0.8cm 0.8cm},clip,height=13mm,width=13mm]{./figure/images/27e8f6a9c6323bd687dc1da2515df8f7_00.png}
        \includegraphics[trim={0.8cm 0.8cm 0.8cm 0.8cm},clip,height=13mm,width=13mm]{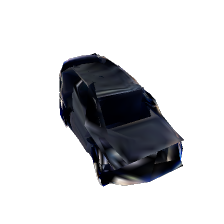}
        \includegraphics[trim={0.8cm 0.8cm 0.8cm 0.8cm},clip,height=13mm,width=13mm]{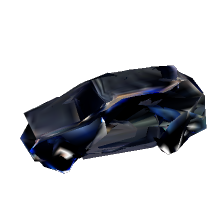}
        \includegraphics[trim={0.8cm 0.8cm 0.8cm 0.8cm},clip,height=13mm,width=13mm]{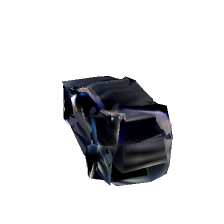}
        \includegraphics[trim={0.8cm 0.8cm 0.8cm 0.8cm},clip,height=13mm,width=13mm]{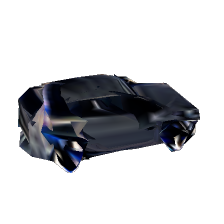} \\
        \raisebox{5.5mm}{\makebox[13mm][l]{\small{Baseline}}}
        \includegraphics[trim={0.8cm 0.8cm 0.8cm 0.8cm},clip,height=13mm,width=13mm]{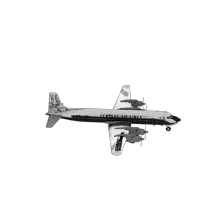}
        \includegraphics[trim={0.8cm 0.8cm 0.8cm 0.8cm},clip,height=13mm,width=13mm]{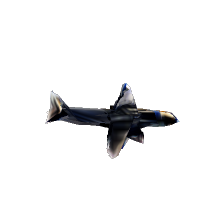}
        \includegraphics[trim={0.8cm 0.8cm 0.8cm 0.8cm},clip,height=13mm,width=13mm]{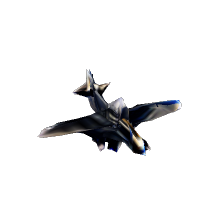}
        \includegraphics[trim={0.8cm 0.8cm 0.8cm 0.8cm},clip,height=13mm,width=13mm]{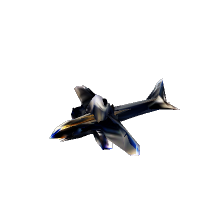}
        \includegraphics[trim={0.8cm 0.8cm 0.8cm 0.8cm},clip,height=13mm,width=13mm]{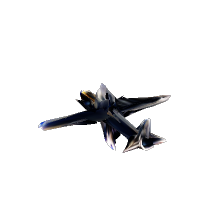} \\
        \raisebox{5.5mm}{\makebox[13mm][l]{\small{Proposed}}}
        \includegraphics[trim={0.8cm 0.8cm 0.8cm 0.8cm},clip,height=13mm,width=13mm]{./figure/images/2ef20ad4c579812571d03b466c72ce41_00.png}
        \includegraphics[trim={0.8cm 0.8cm 0.8cm 0.8cm},clip,height=13mm,width=13mm]{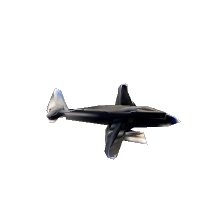}
        \includegraphics[trim={0.8cm 0.8cm 0.8cm 0.8cm},clip,height=13mm,width=13mm]{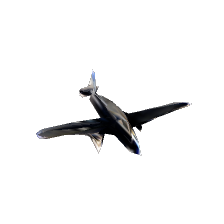}
        \includegraphics[trim={0.8cm 0.8cm 0.8cm 0.8cm},clip,height=13mm,width=13mm]{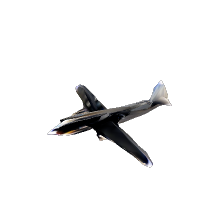}
        \includegraphics[trim={0.8cm 0.8cm 0.8cm 0.8cm},clip,height=13mm,width=13mm]{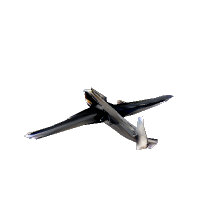} \\
        \vspace{-1mm}
        \makebox[13mm][c]{}
        \makebox[13mm][c]{\small{(a)}}
        \makebox[13mm][c]{\small{(b)}}
        \makebox[13mm][c]{\small{(c)}}
        \makebox[13mm][c]{\small{(d)}}
        \makebox[13mm][c]{\small{(e)}}
        \vspace{-2mm}
    \end{center}
    \caption{Examples of single-view training on the ShapeNet dataset. (a) Input images. (b) Reconstructed shapes viewed from the original viewpoints. (c--e) Reconstructed shapes viewed from other viewpoints.}
    \label{fig:shapenet}
    \vspace{-4mm}
\end{figure}

Figure~\ref{fig:shapenet} shows examples of reconstructed shapes with and without VPL. When viewed from the original viewpoints (b), the estimated shapes appear valid in all cases. However, without VPL, the shapes appear incorrect when viewed from other viewpoints (c--e). For example, the backrest of the chair is too thick, the car is completely broken, and the airplane has a strange prominence in the center. When VPL is used, the shapes look reasonable from any viewpoint. These results clearly indicate that the discriminator informed the reconstructor regarding knowledge of feasible views.

\begin{table*}[t]
    \begin{center}
        \small
        \begin{tabular}{ccc@{\hspace{0.5cm}}ccccccccccccc@{\hspace{0.5cm}}c}
            \toprule
            \STABBa{VPL} & \STABBa{CC} &  \STABBa{TP} & \STABBa{airplane} & \STABBa{bench} & \STABBa{dresser} & \STABBa{car} & \STABBa{chair} & \STABBa{display} & \STABBa{lamp} & \STABBa{speaker} & \STABBa{rifle} & \STABBa{sofa} & \STABBa{table} & \STABBa{phone} & \STABBa{vessel} & \STABBa{all}  \\
            \hline
                       &            &            & $.479$ & $.266$ & $.466$ & $.550$ & $.367$ & $.265$ & $\mathbf{.454}$ & $.524$ & $.382$ & $.367$ & $.342$ & $.337$ & $.439$ & $.403$ \\
            \checkmark &            &            & $.500$ & $.347$ & $.583$ & $.673$ & $.413$ & $.399$ & $.443$ & $.578$ & $.481$ & $.464$ & $.423$ & $.583$ & $.486$ & $.490$ \\
            \checkmark & \checkmark &            & $.513$ & $.376$ & $\mathbf{.591}$ & $.701$ & $.444$ & $\mathbf{.425}$ & $.422$ & $\mathbf{.596}$ & $.479$ & $.500$ & $.436$ & $.595$ & $.485$ & $.505$ \\
                       &            & \checkmark & $.483$ & $.284$ & $.544$ & $.535$ & $.356$ & $.372$ & $.443$ & $.534$ & $.386$ & $.370$ & $.361$ & $.529$ & $.448$ & $.434$ \\
            \checkmark &            & \checkmark & $.524$ & $.378$ & $.581$ & $\mathbf{.705}$ & $.442$ & $.422$ & $.441$ & $.561$ & $.510$ & $.475$ & $.443$ & $\mathbf{.625}$ & $.490$ & $.508$ \\
            \checkmark & \checkmark & \checkmark & $\mathbf{.531}$ & $\mathbf{.385}$ & $\mathbf{.591}$ & $.701$ & $\mathbf{.454}$ & $.423$ & $.441$ & $.570$ & $\mathbf{.521}$ & $\mathbf{.508}$ & $\mathbf{.444}$ & $.601$ & $\mathbf{.498}$ & $\mathbf{.513}$ \\
            \bottomrule
        \end{tabular}
    \end{center}
    \vspace{-1mm}
    \caption{IoU of single-view training on the ShapeNet dataset. VPL: proposed view prior learning. CC: class conditioning in the discriminator. TP: texture prediction. }
    \label{table:shapenet}
    \vspace{-2mm}
\end{table*}

Table~\ref{table:shapenet} shows a quantitative evaluation of single-view training. VPL provides significantly improved reconstruction performance. This improvement is further boosted when the discriminator is class conditioned. We can tell that conducting texture prediction also helps train accurate reconstructors.

VPL is particularly effective with the {\it phone}, {\it display}, {\it bench}, and {\it sofa} categories. In contrast, VPL is not effective with the {\it lamp} category. Typical examples in these categories are shown in the supplementary material. In the case of {\it phone} and {\it display} categories, because the silhouettes are very simple, the shapes are ambiguous and various shapes can fit into one view. Although integrating texture prediction reduces the ambiguity, VPL is much more effective. In the case of {\it bench} and {\it sofa} categories, learning their long shapes is difficult without considering several views. Because the shapes in the {\it lamp} category are diverse and the training dataset is relatively small, the discriminator cannot learn meaningful priors.

%%%%%%%%%%%%%%%%%%%%%%%%%%%%%%%%%%%%%%%%%%%%%%%%%%%%%%%%%%%%%%%%%%%%%%%%%%%%%%%%
\subsubsection{Multi-view training}
\label{sec:exp_multi_view}
\vspace{-1mm}

Second, we trained reconstructors using multi-view training as described in Section~\ref{sec:view_based_training}. Namely, we used two or more views out of twenty views for each object in training.

\begin{figure}[t]
    \begin{center}
        \raisebox{5.5mm}{\makebox[13mm][l]{\small{Baseline}}}
        \includegraphics[trim={0.8cm 0.8cm 0.8cm 0.8cm},clip,height=13mm,width=13mm]{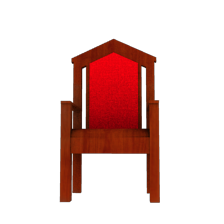}
        \includegraphics[trim={0.8cm 0.8cm 0.8cm 0.8cm},clip,height=13mm,width=13mm]{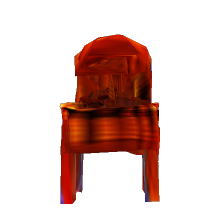}
        \includegraphics[trim={0.8cm 0.8cm 0.8cm 0.8cm},clip,height=13mm,width=13mm]{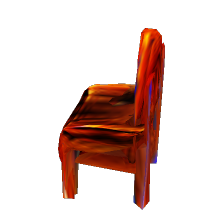}
        \includegraphics[trim={0.8cm 0.8cm 0.8cm 0.8cm},clip,height=13mm,width=13mm]{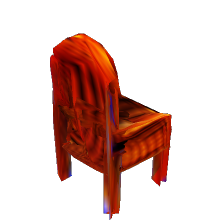}
        \includegraphics[trim={0.8cm 0.8cm 0.8cm 0.8cm},clip,height=13mm,width=13mm]{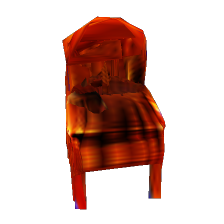} \\
        \raisebox{5.5mm}{\makebox[13mm][l]{\small{Proposed}}}
        \includegraphics[trim={0.8cm 0.8cm 0.8cm 0.8cm},clip,height=13mm,width=13mm]{./figure/images/717e28c855c935c94d2d89cc1fd36fca_00.png}
        \includegraphics[trim={0.8cm 0.8cm 0.8cm 0.8cm},clip,height=13mm,width=13mm]{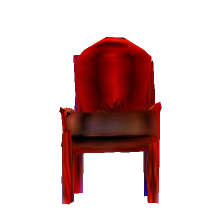}
        \includegraphics[trim={0.8cm 0.8cm 0.8cm 0.8cm},clip,height=13mm,width=13mm]{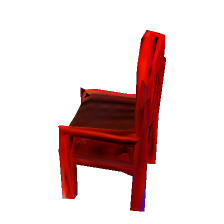}
        \includegraphics[trim={0.8cm 0.8cm 0.8cm 0.8cm},clip,height=13mm,width=13mm]{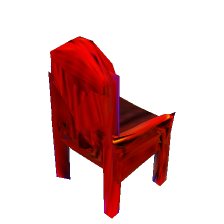}
        \includegraphics[trim={0.8cm 0.8cm 0.8cm 0.8cm},clip,height=13mm,width=13mm]{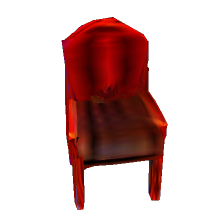} \\
        \raisebox{5.5mm}{\makebox[13mm][l]{\small{Baseline}}}
        \includegraphics[trim={0.8cm 0.8cm 0.8cm 0.8cm},clip,height=13mm,width=13mm]{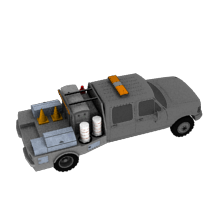}
        \includegraphics[trim={0.8cm 0.8cm 0.8cm 0.8cm},clip,height=13mm,width=13mm]{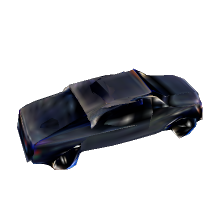}
        \includegraphics[trim={0.8cm 0.8cm 0.8cm 0.8cm},clip,height=13mm,width=13mm]{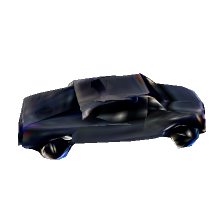}
        \includegraphics[trim={0.8cm 0.8cm 0.8cm 0.8cm},clip,height=13mm,width=13mm]{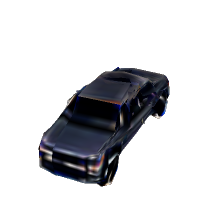}
        \includegraphics[trim={0.8cm 0.8cm 0.8cm 0.8cm},clip,height=13mm,width=13mm]{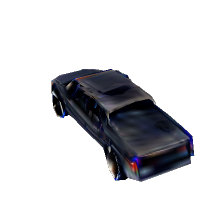} \\
        \raisebox{5.5mm}{\makebox[13mm][l]{\small{Proposed}}}
        \includegraphics[trim={0.8cm 0.8cm 0.8cm 0.8cm},clip,height=13mm,width=13mm]{./figure/images/764d4cd02b799a2b59139efcde1fedcb_00.png}
        \includegraphics[trim={0.8cm 0.8cm 0.8cm 0.8cm},clip,height=13mm,width=13mm]{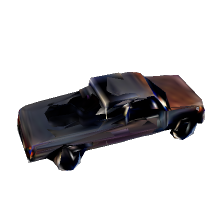}
        \includegraphics[trim={0.8cm 0.8cm 0.8cm 0.8cm},clip,height=13mm,width=13mm]{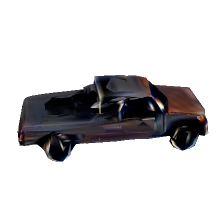}
        \includegraphics[trim={0.8cm 0.8cm 0.8cm 0.8cm},clip,height=13mm,width=13mm]{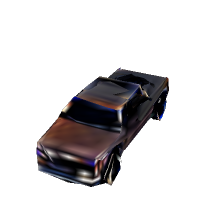}
        \includegraphics[trim={0.8cm 0.8cm 0.8cm 0.8cm},clip,height=13mm,width=13mm]{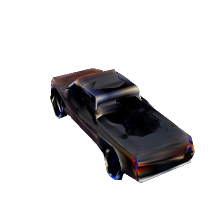} \\
        \raisebox{5.5mm}{\makebox[13mm][l]{\small{Baseline}}}
        \includegraphics[trim={0.8cm 0.8cm 0.8cm 0.8cm},clip,height=13mm,width=13mm]{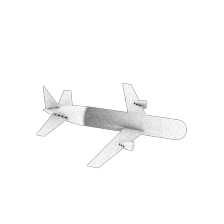}
        \includegraphics[trim={0.8cm 0.8cm 0.8cm 0.8cm},clip,height=13mm,width=13mm]{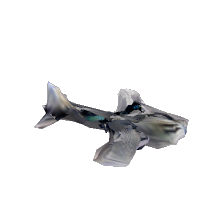}
        \includegraphics[trim={0.8cm 0.8cm 0.8cm 0.8cm},clip,height=13mm,width=13mm]{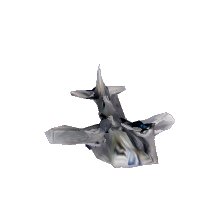}
        \includegraphics[trim={0.8cm 0.8cm 0.8cm 0.8cm},clip,height=13mm,width=13mm]{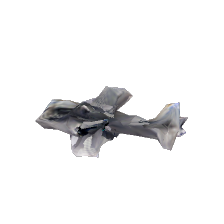}
        \includegraphics[trim={0.8cm 0.8cm 0.8cm 0.8cm},clip,height=13mm,width=13mm]{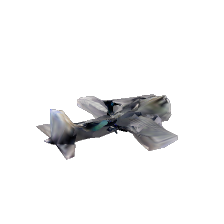} \\
        \raisebox{5.5mm}{\makebox[13mm][l]{\small{Proposed}}}
        \includegraphics[trim={0.8cm 0.8cm 0.8cm 0.8cm},clip,height=13mm,width=13mm]{./figure/images/c9be9f07f5ae7c375d7629390efe0a2_00.png}
        \includegraphics[trim={0.8cm 0.8cm 0.8cm 0.8cm},clip,height=13mm,width=13mm]{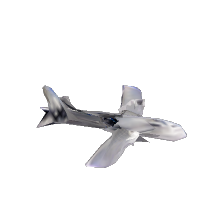}
        \includegraphics[trim={0.8cm 0.8cm 0.8cm 0.8cm},clip,height=13mm,width=13mm]{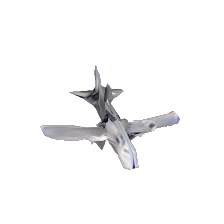}
        \includegraphics[trim={0.8cm 0.8cm 0.8cm 0.8cm},clip,height=13mm,width=13mm]{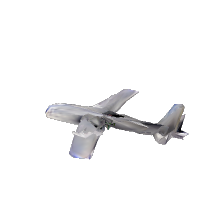}
        \includegraphics[trim={0.8cm 0.8cm 0.8cm 0.8cm},clip,height=13mm,width=13mm]{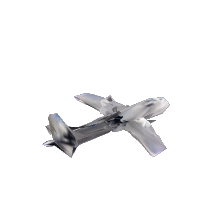} \\
        \vspace{-1mm}
        \makebox[13mm][c]{}
        \makebox[13mm][c]{\small{(a)}}
        \makebox[13mm][c]{\small{(b)}}
        \makebox[13mm][c]{\small{(c)}}
        \makebox[13mm][c]{\small{(d)}}
        \makebox[13mm][c]{\small{(e)}}
        \vspace{-2mm}
    \end{center}
    \caption{Examples of multi-view training on ShapeNet ($N_v = 2$). Panels (a--e) are the same as in Figure~\ref{fig:shapenet}. }
    \label{fig:shapenet_multi_nv2}
\end{figure}

\begin{table}[t]
    \small
    \begin{center}
        \begin{tabular}{lccccc}
            \toprule
            $N_v$      & $2$     & $3$    & $5$    & $10$ & $20$   \\
            \hline
            Baseline   & $.575$  & $.596$ & $.620$ & $.641$ & $.652$ \\
            Proposed   & $\mathbf{.583}$  & $\mathbf{.600}$ & $\mathbf{.624}$ & $\mathbf{.644}$ & $\mathbf{.655}$ \\
            \bottomrule
        \end{tabular}
    \end{center}
    \vspace{-1mm}
    \caption{The relation between the number of views per object $N_v$ and the reconstruction accuracy (IoU) in multi-view training.}
    \label{table:shapenet_views}
\end{table}

Table~\ref{table:shapenet_views} shows the relationship between the reconstruction accuracy and the number of views per object $N_v$ used for training. Texture prediction was not conducted in this experiment, and the difference between the proposed method and the baseline is the use of VPL with class conditioning. Our proposed method outperforms the baseline in all cases, which indicates that VPL is also effective in multi-view training. The effect of VPL increases as $N_v$ decreases, as expected. Figure~\ref{fig:shapenet_multi_nv2} shows reconstructed shapes with texture prediction when $N_v=2$. When VPL is used, the shape details become more accurate.

%%%%%%%%%%%%%%%%%%%%%%%%%%%%%%%%%%%%%%%%%%%%%%%%%%%%%%%%%%%%%%%%%%%%%%%%%%%%%%%%
\subsubsection{Discriminator and optimization}
\label{sec:exp_dis_opt}
\vspace{-1mm}

We discussed two types of discriminators in the last paragraph of Section~\ref{sec:view_prior_learning} and emphasized the importance of discriminating between estimated views rather than estimated views and real views. We validated this statement with an experiment. We ran experiments in single-view training using the discriminator of Table~\ref{table:discriminators} (d). We also tested the iterative optimization used in GAN~\cite{goodfellow2014generative} instead of using a gradient reversal layer~\cite{ganin2016domain}. However, in both cases, we were unable to observe any meaningful improvements from the baseline by tuning $\lambda_d$. This fact indicates that the discriminator in Figure~\ref{table:discriminators} (d) does not work well in practice, and discriminating estimated views is key to effective training.

%%%%%%%%%%%%%%%%%%%%%%%%%%%%%%%%%%%%%%%%%%%%%%%%%%%%%%%%%%%%%%%%%%%%%%%%%%%%%%%%
\subsubsection{Comparison with manually-designed priors}
\label{sec:exp_priors}
\vspace{-1mm}

\begin{table}[t]
    \small
    \begin{center}
        \begin{tabular}{lc}
            \toprule
            Prior                                                                              & IoU    \\
            \hline
            None                                                                               & $.387$ \\
            Internal pressure (IP, ours)                                                       & $.403$ \\
            IP \& Symmetricity~\cite{kanazawa2018learning}                                     & $.420$ \\
            IP \& Regularizing graph Laplacian~\cite{kanazawa2018learning,wang2018pixel2mesh}  & $\mathit{.403}^*$ \\
            IP \& Regularizing edge length~\cite{wang2018pixel2mesh}                           & $\mathit{.403}^*$ \\
            IP \& View prior learning (ours)                                                   & $\mathbf{.505}$ \\
            \bottomrule
        \end{tabular}
    \end{center}
    \vspace{-1mm}
    \caption{Comparison of our learning-based prior with manually-designed shape regularizers and constraints. ${}^*$No meaningful improvement was observed.}
    \label{table:regularizer}
\end{table}

Our proposed internal pressure (IP) loss and some regularizers and constraints used in~\cite{kanazawa2018learning,wang2018pixel2mesh} were designed using human knowledge regarding shapes. Table~\ref{table:regularizer} shows a comparison with VPL. This experiment was conducted in single-view training without texture prediction.

This result shows that IP loss improves performance. The symmetricity constraint also improves the performance, however, some objects in ShapeNet are actually not symmetric. By regularizing the graph Laplacian and the edge length of meshes, although the visual quality of the generated meshes became better, improvement of IoU was not observed.

VPL cannot be compared with the learning-based 3D shape priors detailed by Gwak \etal~\cite{gwak2017weakly} and Wu \etal~\cite{wu2018learning} because these methods require additional 3D models for training, and their methods are applicable to voxels rather than meshes.

%%%%%%%%%%%%%%%%%%%%%%%%%%%%%%%%%%%%%%%%%%%%%%%%%%%%%%%%%%%%%%%%%%%%%%%%%%%%%%%%
\subsubsection{Comparison with state-of-the-arts}
\vspace{-1mm}

\begin{table}[t]
    \small
    \begin{center}
        \begin{tabular}{lcc}
            \toprule
            & $N_v$ & IoU  \\
            \hline
            Single-view training \\
            Our best model${}^\sharp$                              & $1$  & $\mathbf{.513}$ \\
            \hline
            Multi-view training \\
            PTN~\cite{yan2016perspective}                        & $24$ & $.574$ \\
            NMR~\cite{kato2018neural}                            & $24$ & $.602$ \\
            Our best model${}^\sharp$                              & $20$ & $\mathbf{.655}$ \\
            \hline
            3D supervision \\
            3D-R2N2${}^\sharp$~\cite{kar2017learning}              & $20$ & $.551$ \\
            3D-R2N2${}^\flat$~\cite{choy20163d}                  & $24$ & $.560$ \\
            OGN${}^\flat$~\cite{tatarchenko2017octree}           & $24$ & $.596$ \\
            LSM${}^\sharp$~\cite{kar2017learning}                  & $20$ & $.615$ \\
            Matryoshka${}^\flat$~\cite{richter2018matryoshka}    & $24$ & $.635$ \\
            PSGN${}^\flat$~\cite{fan2016point}                   & $24$ & $.640$ \\
            VTN${}^\flat$~\cite{richter2018matryoshka}           & $24$ & $\mathbf{.641}$ \\
            \bottomrule
        \end{tabular}
    \end{center}
    \vspace{-1mm}
    \caption{Comparison of our method and state-of-the-art methods on ShapeNet (3D-R2N2) dataset using IoU. Although supervision is weaker, our proposed method outperforms the other models trained using 3D models. ${}^{\sharp\flat}$Models denoted with the same symbol use the same rendered images.}
    \label{table:shapenet_sota}
\end{table}

Our work also shows the effectiveness of view-based training. Table~\ref{table:shapenet_sota} shows the reconstruction accuracy (IoU) on the ShapeNet dataset using our method and that presented in recent papers\footnote{The most commonly used dataset of ShapeNet for 3D reconstruction was provided by Choy \etal~\cite{choy20163d}. However, we found that this dataset is not suitable for view-based training because there are large occluded regions in the views owing to the narrow range of elevation in the viewpoints. Therefore, we used a dataset by Kar \etal~\cite{kar2017learning}, in which images were rendered from a variety of viewpoints. A comparison of the results from both datasets is not so unfair because the performance of 3D-R2N2~\cite{choy20163d} is close in both datasets.}\footnote{This table only compares papers that report IoU on 3D-R2N2 dataset. On other metrics and datasets, some works~\cite{groueix2018atlasnet,wang2018pixel2mesh} outperform PSGN~\cite{fan2016point}.}. Our method outperforms existing view-based training methods~\cite{kato2018neural,yan2016perspective}. The main differences between our baseline and~\cite{kato2018neural} are the internal pressure and the training dataset. Because the resolution of our training images ($224 \times 224$) is larger than theirs ($64 \times 64$) and the elevation range in the viewpoints ($[-20^\circ, 30^\circ]$) is wider than that of theirs ($30^\circ$ only), more accurate and detailed 3D shapes can be learned in our experiments.

It may be surprising that our view-based method outperforms reconstructors trained using 3D models. Although view-based training is currently less popular than 3D-based training, one can say that view-based training has much room for further study.

%%%%%%%%%%%%%%%%%%%%%%%%%%%%%%%%%%%%%%%%%%%%%%%%%%%%%%%%%%%%%%%%%%%%%%%%%%%%%%%%
\subsection{Natural image dataset}
\vspace{-1mm}

If a 3D model is available, we can synthetically create multiple views with accurate silhouette and viewpoint annotations. However, in practical applications, it is not always possible to obtain many 3D models, and datasets must be created using natural images. In this case, generally, multi-view training is not possible, and silhouette and viewpoint annotations are noisy. Therefore, to measure the practicality of a given method, it is important to evaluate such a case.

Thus, we used the PASCAL dataset preprocessed by Tulsiani \etal~\cite{tulsiani2017multi}. This dataset is composed of images in PASCAL VOC~\cite{everingham2010pascal}, annotations of 3D models, silhouettes, and viewpoints in PASCAL 3D+~\cite{xiang2014beyond}, and additional images in ImageNet~\cite{russakovsky2015imagenet} with silhouette and viewpoint annotations automatically created using~\cite{li2016iterative}. We conducted single-view training because there is only one view per object. Because this dataset is not large, the variance in the training results is not negligible. Therefore, we report the mean accuracy from five runs with different random seeds. We used the pre-trained ResNet-18 model~\cite{he2016deep} as the encoder as with~\cite{kanazawa2018learning,tulsiani2017multi}. The parameters were optimized with the Adam optimizer~\cite{kingma2014adam}. The architecture of decoders, discriminators, and other hyperparameters are described in the supplementary material. We constrained estimated shapes to be symmetric, as was the case in a previous study~\cite{kanazawa2018learning}. We used Equation~\ref{eq:silhouette2} as the view comparison function for silhouettes.

\begin{table}[t]
    \small
    \begin{center}
        \begin{tabular}{l@{\hspace{0.5cm}}ccc@{\hspace{0.5cm}}c}
            \toprule
            & airplane & car & chair & mean \\
            \hline
            \multicolumn{1}{l}{Category-agnostic models}\\
            DRC~\cite{tulsiani2017multi}    & $.415$ & $.666$ & $.247$ & $.443$ \\
            Baseline (s)                    & $.448$ & $.652$ & $.272$ & $.458$ \\
            Proposed (s)                    & $.450$ & $\mathbf{.672}$ & $.292$ & $.471$ \\
            Baseline                        & $.440$ & $.640$ & $.280$ & $.454$ \\
            Proposed                        & $\mathbf{.460}$ & $.662$ & $\mathbf{.296}$ & $\mathbf{.473}$ \\
            \hline
            \multicolumn{1}{l}{Category-specific models} \\
            CSDM~\cite{kar2015category}     & $.398$ & $.600$ & $.291$ & $.429$ \\
            CMR~\cite{kanazawa2018learning} & $.46$  & $.64$  & n/a    & n/a    \\
            Baseline (s)                    & $.449$ & $.679$ & $.289$ & $.472$ \\
            Proposed (s)                    & $.472$ & $\mathbf{.689}$ & $.303$ & $\mathbf{.488}$ \\
            Baseline                        & $.450$ & $.669$ & $.293$ & $.470$ \\
            Proposed                        & $\mathbf{.475}$ & $.679$ & $\mathbf{.304}$ & $.486$ \\
            \bottomrule
        \end{tabular}
    \end{center}
    \vspace{-1mm}
    \caption{IoU of single-view 3D reconstruction on the PASCAL dataset. The difference between the proposed method and the baseline is the use of view prior learning. (s) indicates silhouette only training without texture prediction ($\lambda_c = 0$).}
    \label{table:pascal}
    \vspace{-2mm}
\end{table}

Table~\ref{table:pascal} shows the reconstruction accuracy on the PASCAL dataset. Our proposed method consistently outperforms the baseline and provides state-of-the-art performance for this dataset, which validates the effectiveness of our proposed method. Category-specific models outperform category-agnostic models because the object shapes in these three categories are not very similar and multitask learning is not beneficial. The performance difference when texture prediction is used is primarily caused by the relative weight of the internal pressure loss.

\begin{figure}[t]
    \begin{center}
        \raisebox{5.5mm}{\makebox[13mm][l]{\small{Baseline}}}
        \includegraphics[height=13mm,width=13mm]{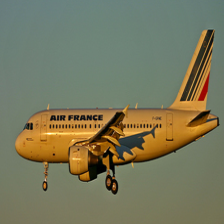}
        \includegraphics[trim={0.6cm 0.6cm 0.6cm 0.6cm},clip,height=13mm,width=13mm]{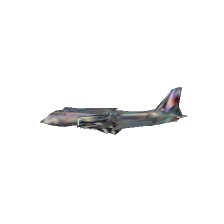}
        \includegraphics[trim={0.6cm 0.6cm 0.6cm 0.6cm},clip,height=13mm,width=13mm]{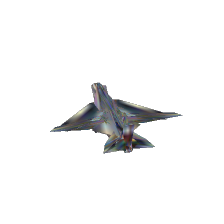}
        \includegraphics[trim={0.6cm 0.6cm 0.6cm 0.6cm},clip,height=13mm,width=13mm]{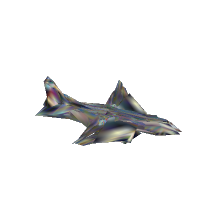}
        \includegraphics[trim={0.6cm 0.6cm 0.6cm 0.6cm},clip,height=13mm,width=13mm]{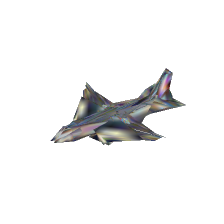}
        \raisebox{5.5mm}{\makebox[13mm][l]{\small{Proposed}}}
        \includegraphics[height=13mm,width=13mm]{./figure/images/pascal_aeroplane_02.png}
        \includegraphics[trim={0.6cm 0.6cm 0.6cm 0.6cm},clip,height=13mm,width=13mm]{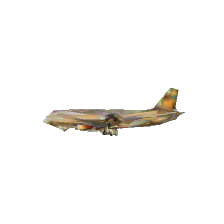}
        \includegraphics[trim={0.6cm 0.6cm 0.6cm 0.6cm},clip,height=13mm,width=13mm]{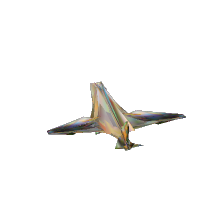}
        \includegraphics[trim={0.6cm 0.6cm 0.6cm 0.6cm},clip,height=13mm,width=13mm]{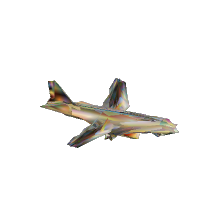}
        \includegraphics[trim={0.6cm 0.6cm 0.6cm 0.6cm},clip,height=13mm,width=13mm]{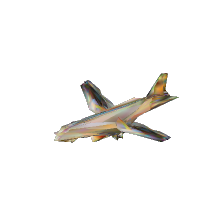}
        \raisebox{5.5mm}{\makebox[13mm][l]{\small{Baseline}}}
        \includegraphics[height=13mm,width=13mm]{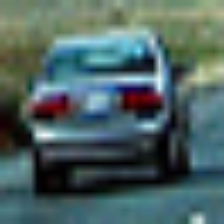}
        \includegraphics[trim={0.6cm 0.6cm 0.6cm 0.6cm},clip,height=13mm,width=13mm]{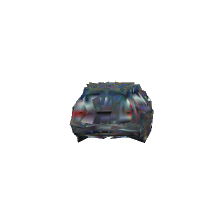}
        \includegraphics[trim={0.6cm 0.6cm 0.6cm 0.6cm},clip,height=13mm,width=13mm]{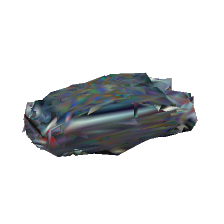}
        \includegraphics[trim={0.6cm 0.6cm 0.6cm 0.6cm},clip,height=13mm,width=13mm]{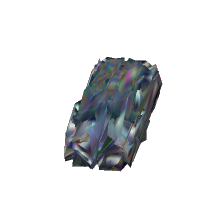}
        \includegraphics[trim={0.6cm 0.6cm 0.6cm 0.6cm},clip,height=13mm,width=13mm]{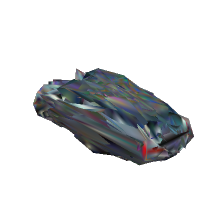}
        \raisebox{5.5mm}{\makebox[13mm][l]{\small{Proposed}}}
        \includegraphics[height=13mm,width=13mm]{./figure/images/pascal_car_10.png}
        \includegraphics[trim={0.6cm 0.6cm 0.6cm 0.6cm},clip,height=13mm,width=13mm]{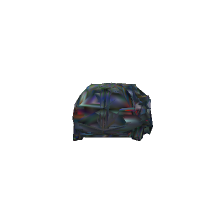}
        \includegraphics[trim={0.6cm 0.6cm 0.6cm 0.6cm},clip,height=13mm,width=13mm]{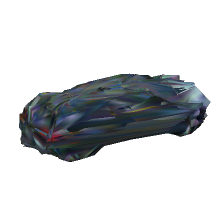}
        \includegraphics[trim={0.6cm 0.6cm 0.6cm 0.6cm},clip,height=13mm,width=13mm]{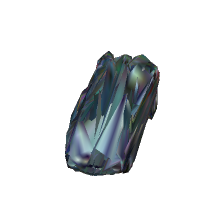}
        \includegraphics[trim={0.6cm 0.6cm 0.6cm 0.6cm},clip,height=13mm,width=13mm]{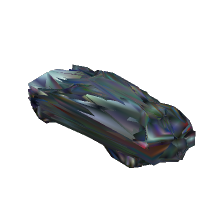}
        \raisebox{5.5mm}{\makebox[13mm][l]{\small{Baseline}}}
        \includegraphics[height=13mm,width=13mm]{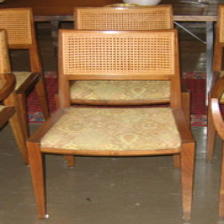}
        \includegraphics[trim={0.6cm 0.6cm 0.6cm 0.6cm},clip,height=13mm,width=13mm]{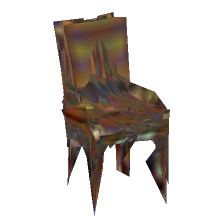}
        \includegraphics[trim={0.6cm 0.6cm 0.6cm 0.6cm},clip,height=13mm,width=13mm]{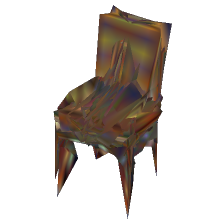}
        \includegraphics[trim={0.6cm 0.6cm 0.6cm 0.6cm},clip,height=13mm,width=13mm]{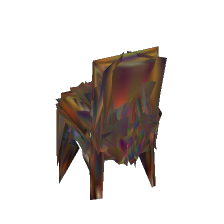}
        \includegraphics[trim={0.6cm 0.6cm 0.6cm 0.6cm},clip,height=13mm,width=13mm]{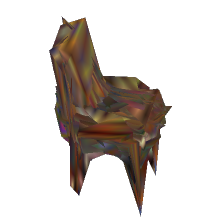}
        \raisebox{5.5mm}{\makebox[13mm][l]{\small{Proposed}}}
        \includegraphics[height=13mm,width=13mm]{./figure/images/pascal_chair_09.png}
        \includegraphics[trim={0.6cm 0.6cm 0.6cm 0.6cm},clip,height=13mm,width=13mm]{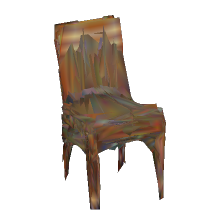}
        \includegraphics[trim={0.6cm 0.6cm 0.6cm 0.6cm},clip,height=13mm,width=13mm]{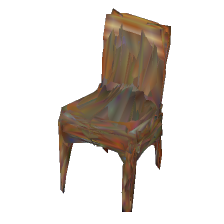}
        \includegraphics[trim={0.6cm 0.6cm 0.6cm 0.6cm},clip,height=13mm,width=13mm]{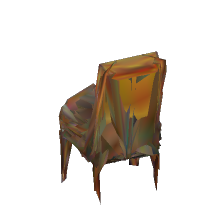}
        \includegraphics[trim={0.6cm 0.6cm 0.6cm 0.6cm},clip,height=13mm,width=13mm]{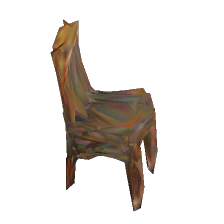} \\
        \vspace{-1mm}
        \makebox[13mm][c]{}
        \makebox[13mm][c]{\small{(a)}}
        \makebox[13mm][c]{\small{(b)}}
        \makebox[13mm][c]{\small{(c)}}
        \makebox[13mm][c]{\small{(d)}}
        \makebox[13mm][c]{\small{(e)}}
        \vspace{-2mm}
    \end{center}
    \caption{Examples on the PASCAL dataset. Panels (a--e) are the same as in Figure~\ref{fig:shapenet}.}
    \label{fig:pascal}
    \vspace{-1mm}
\end{figure}

\begin{figure}[t]
    \begin{center}
        \raisebox{7.0mm}{\makebox[13mm][l]{\parbox[t]{13mm}{\small{Proposed\\w/o IP}}}}
        \includegraphics[height=13mm,width=13mm]{./figure/images/pascal_car_10.png}
        \includegraphics[trim={0.6cm 0.6cm 0.6cm 0.6cm},clip,height=13mm,width=13mm]{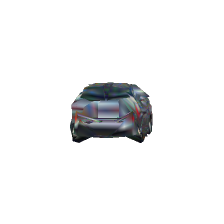}
        \includegraphics[trim={0.6cm 0.6cm 0.6cm 0.6cm},clip,height=13mm,width=13mm]{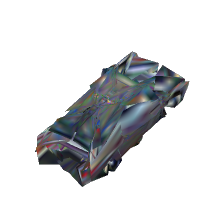}
        \includegraphics[trim={0.6cm 0.6cm 0.6cm 0.6cm},clip,height=13mm,width=13mm]{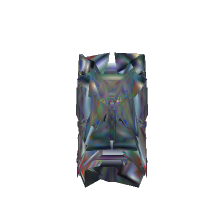}
        \includegraphics[trim={0.6cm 0.6cm 0.6cm 0.6cm},clip,height=13mm,width=13mm]{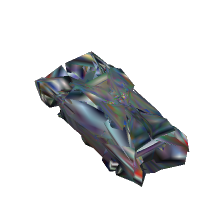} \\
        \vspace{-1mm}
        \makebox[13mm][c]{}
        \makebox[13mm][c]{\small{(a)}}
        \makebox[13mm][c]{\small{(b)}}
        \makebox[13mm][c]{\small{(c)}}
        \makebox[13mm][c]{\small{(d)}}
        \makebox[13mm][c]{\small{(e)}}
        \vspace{-2mm}
    \end{center}
    \caption{An example of reconstruction without internal pressure (IP). Panels (a--e) are the same as in Figure~\ref{fig:shapenet}.}
    \label{fig:pascal_wo_ip}
    \vspace{-2mm}
\end{figure}

Figure~\ref{fig:pascal} shows typical improvements that can be gained using our method. Improvements are prominent on the wings of the airplane, the tires of the car, and the front legs of the chair when viewed from unobserved viewpoints.

In this experiment, internal pressure loss plays an important role because observed viewpoints are not diverse. Figure~\ref{fig:pascal_wo_ip} shows a reconstructed shape without internal pressure. The trunk of the car is hollowed, and this hollow cannot be filled by VPL because there are few images taken from viewpoints such as (c--e) in the dataset.

\section{Conclusion}
\vspace{-1mm}

In this work, we proposed a method to learn prior knowledge of views for view-based training of 3D object reconstructors. We verified our approach in single-view training on both synthetic and natural image datasets. We also found that our method is effective, even when multiple views are available for training. The key to our success involves using a discriminator with two estimated views from observed and unobserved viewpoints. Our data-driven method works better than existing manually-designed shape regularizers. We also showed that view-based training works as well as methods that use 3D models for training. The experimental results clearly validate these statements.

Our method significantly improves reconstruction accuracy, especially in single-view training. This is important progress because it is easier to create a single-view dataset than to create a multi-view dataset. This fact may enable 3D reconstruction of diverse objects beyond the existing synthetic datasets. The most important limitation of our method is that it requires silhouette and viewpoint annotations. Training end-to-end 3D reconstruction, viewpoint prediction, and silhouette segmentation would be a promising future direction.

\subsection*{Acknowledgment}
\addcontentsline{toc}{section}{Acknowledgment}
\vspace{-1mm}

\small{This work was partially funded by ImPACT Program of Council for Science, Technology and Innovation (Cabinet Office, Government of Japan) and partially supported by JST CREST Grant Number JPMJCR1403, Japan. We would like to thank Antonio Tejero de Pablos, Atsuhiro Noguchi, Kosuke Arase, and Takuhiro Kaneko for helpful discussions.}

%-------------------------------------------------------------------------
\clearpage
{
    \small
    \bibliographystyle{ieee}
    \bibliography{paper}
}

%-------------------------------------------------------------------------
\clearpage
\appendix
%%%%%%%%%%%%%%%%%%%%%%%%%%%%%%%%%%%%%%%%%%%%%%%%%%%%%%%%%%%%%%%%%%%%%%%%%%%%%%%%
\section{Appendix}

%%%%%%%%%%%%%%%%%%%%%%%%%%%%%%%%%%%%%%%%%%%%%%%%%%%%%%%%%%%%%%%%%%%%%%%%%%%%%%%%
\subsection{Additional examples of single-view training}

\begin{figure}[!t]
    \begin{center}
        \raisebox{5.5mm}{\makebox[13mm][l]{\small{Baseline}}}
        \includegraphics[trim={0.8cm 0.8cm 0.8cm 0.8cm},clip,height=13mm,width=13mm]{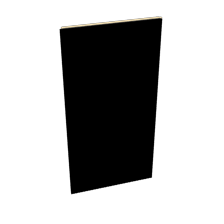}
        \includegraphics[trim={0.8cm 0.8cm 0.8cm 0.8cm},clip,height=13mm,width=13mm]{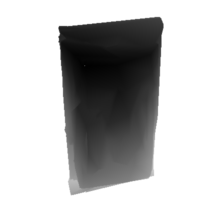}
        \includegraphics[trim={0.8cm 0.8cm 0.8cm 0.8cm},clip,height=13mm,width=13mm]{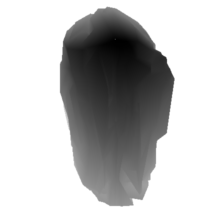}
        \includegraphics[trim={0.8cm 0.8cm 0.8cm 0.8cm},clip,height=13mm,width=13mm]{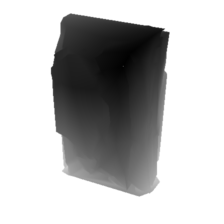}
        \includegraphics[trim={0.8cm 0.8cm 0.8cm 0.8cm},clip,height=13mm,width=13mm]{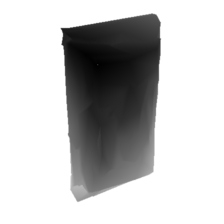} \\
        \raisebox{5.5mm}{\makebox[13mm][l]{\small{Proposed}}}
        \includegraphics[trim={0.8cm 0.8cm 0.8cm 0.8cm},clip,height=13mm,width=13mm]{./figure/images/112cdf6f3466e35fa36266c295c27a25_00.png}
        \includegraphics[trim={0.8cm 0.8cm 0.8cm 0.8cm},clip,height=13mm,width=13mm]{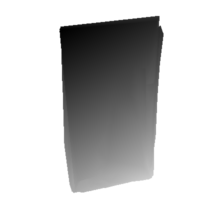}
        \includegraphics[trim={0.8cm 0.8cm 0.8cm 0.8cm},clip,height=13mm,width=13mm]{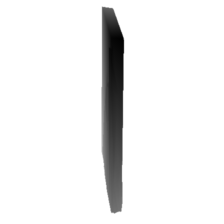}
        \includegraphics[trim={0.8cm 0.8cm 0.8cm 0.8cm},clip,height=13mm,width=13mm]{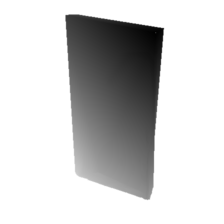}
        \includegraphics[trim={0.8cm 0.8cm 0.8cm 0.8cm},clip,height=13mm,width=13mm]{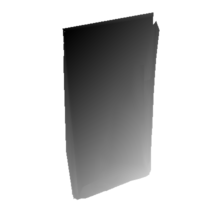} \\
        \raisebox{5.5mm}{\makebox[13mm][l]{\small{Baseline}}}
        \includegraphics[trim={0.8cm 0.8cm 0.8cm 0.8cm},clip,height=13mm,width=13mm]{./figure/images/112cdf6f3466e35fa36266c295c27a25_00.png}
        \includegraphics[trim={0.8cm 0.8cm 0.8cm 0.8cm},clip,height=13mm,width=13mm]{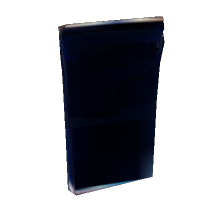}
        \includegraphics[trim={0.8cm 0.8cm 0.8cm 0.8cm},clip,height=13mm,width=13mm]{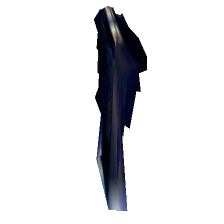}
        \includegraphics[trim={0.8cm 0.8cm 0.8cm 0.8cm},clip,height=13mm,width=13mm]{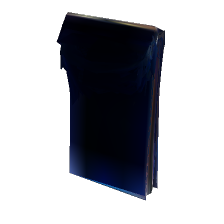}
        \includegraphics[trim={0.8cm 0.8cm 0.8cm 0.8cm},clip,height=13mm,width=13mm]{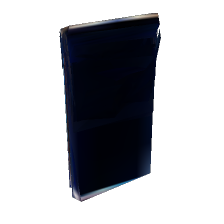} \\
        \raisebox{5.5mm}{\makebox[13mm][l]{\small{Proposed}}}
        \includegraphics[trim={0.8cm 0.8cm 0.8cm 0.8cm},clip,height=13mm,width=13mm]{./figure/images/112cdf6f3466e35fa36266c295c27a25_00.png}
        \includegraphics[trim={0.8cm 0.8cm 0.8cm 0.8cm},clip,height=13mm,width=13mm]{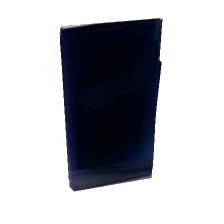}
        \includegraphics[trim={0.8cm 0.8cm 0.8cm 0.8cm},clip,height=13mm,width=13mm]{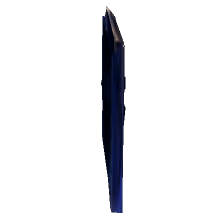}
        \includegraphics[trim={0.8cm 0.8cm 0.8cm 0.8cm},clip,height=13mm,width=13mm]{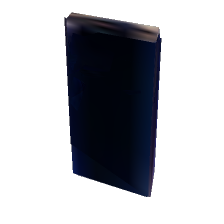}
        \includegraphics[trim={0.8cm 0.8cm 0.8cm 0.8cm},clip,height=13mm,width=13mm]{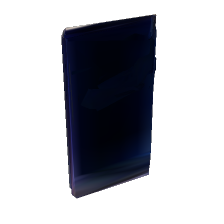} \\
        \raisebox{5.5mm}{\makebox[13mm][l]{\small{Baseline}}}
        \includegraphics[trim={0.8cm 0.8cm 0.8cm 0.8cm},clip,height=13mm,width=13mm]{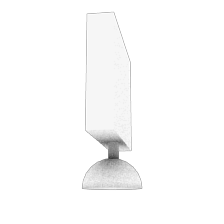}
        \includegraphics[trim={0.8cm 0.8cm 0.8cm 0.8cm},clip,height=13mm,width=13mm]{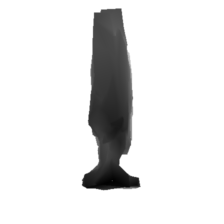}
        \includegraphics[trim={0.8cm 0.8cm 0.8cm 0.8cm},clip,height=13mm,width=13mm]{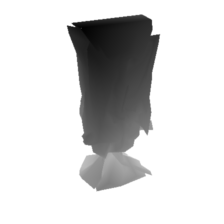}
        \includegraphics[trim={0.8cm 0.8cm 0.8cm 0.8cm},clip,height=13mm,width=13mm]{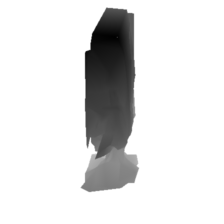}
        \includegraphics[trim={0.8cm 0.8cm 0.8cm 0.8cm},clip,height=13mm,width=13mm]{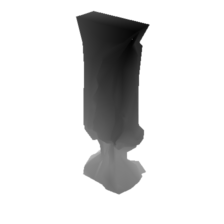} \\
        \raisebox{5.5mm}{\makebox[13mm][l]{\small{Proposed}}}
        \includegraphics[trim={0.8cm 0.8cm 0.8cm 0.8cm},clip,height=13mm,width=13mm]{./figure/images/538eebd970e4870546ed33fa3575cd87_00.png}
        \includegraphics[trim={0.8cm 0.8cm 0.8cm 0.8cm},clip,height=13mm,width=13mm]{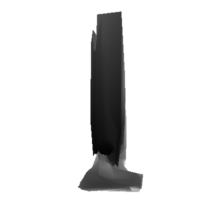}
        \includegraphics[trim={0.8cm 0.8cm 0.8cm 0.8cm},clip,height=13mm,width=13mm]{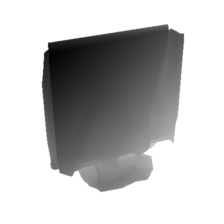}
        \includegraphics[trim={0.8cm 0.8cm 0.8cm 0.8cm},clip,height=13mm,width=13mm]{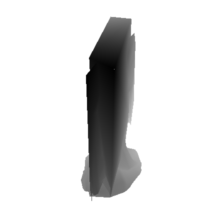}
        \includegraphics[trim={0.8cm 0.8cm 0.8cm 0.8cm},clip,height=13mm,width=13mm]{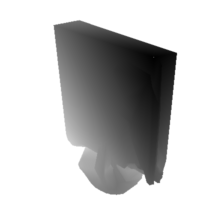} \\
        \raisebox{5.5mm}{\makebox[13mm][l]{\small{Baseline}}}
        \includegraphics[trim={0.8cm 0.8cm 0.8cm 0.8cm},clip,height=13mm,width=13mm]{./figure/images/538eebd970e4870546ed33fa3575cd87_00.png}
        \includegraphics[trim={0.8cm 0.8cm 0.8cm 0.8cm},clip,height=13mm,width=13mm]{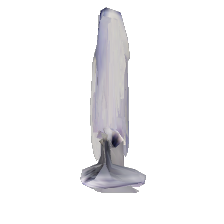}
        \includegraphics[trim={0.8cm 0.8cm 0.8cm 0.8cm},clip,height=13mm,width=13mm]{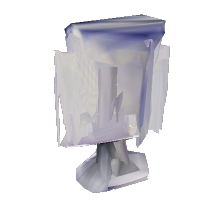}
        \includegraphics[trim={0.8cm 0.8cm 0.8cm 0.8cm},clip,height=13mm,width=13mm]{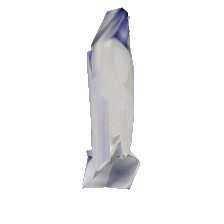}
        \includegraphics[trim={0.8cm 0.8cm 0.8cm 0.8cm},clip,height=13mm,width=13mm]{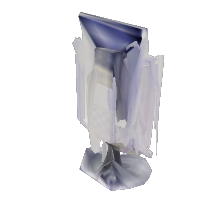} \\
        \raisebox{5.5mm}{\makebox[13mm][l]{\small{Proposed}}}
        \includegraphics[trim={0.8cm 0.8cm 0.8cm 0.8cm},clip,height=13mm,width=13mm]{./figure/images/538eebd970e4870546ed33fa3575cd87_00.png}
        \includegraphics[trim={0.8cm 0.8cm 0.8cm 0.8cm},clip,height=13mm,width=13mm]{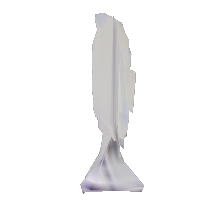}
        \includegraphics[trim={0.8cm 0.8cm 0.8cm 0.8cm},clip,height=13mm,width=13mm]{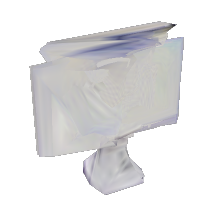}
        \includegraphics[trim={0.8cm 0.8cm 0.8cm 0.8cm},clip,height=13mm,width=13mm]{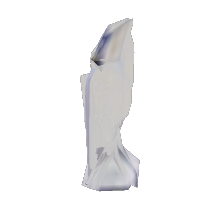}
        \includegraphics[trim={0.8cm 0.8cm 0.8cm 0.8cm},clip,height=13mm,width=13mm]{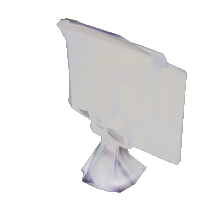} \\
        \raisebox{5.5mm}{\makebox[13mm][l]{\small{Baseline}}}
        \includegraphics[trim={0.8cm 0.8cm 0.8cm 0.8cm},clip,height=13mm,width=13mm]{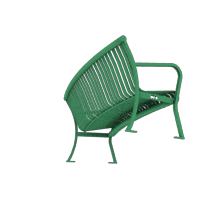}
        \includegraphics[trim={0.8cm 0.8cm 0.8cm 0.8cm},clip,height=13mm,width=13mm]{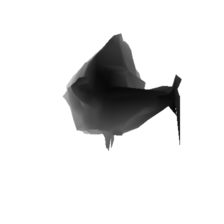}
        \includegraphics[trim={0.8cm 0.8cm 0.8cm 0.8cm},clip,height=13mm,width=13mm]{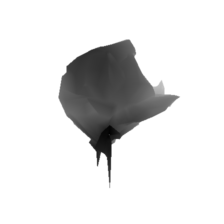}
        \includegraphics[trim={0.8cm 0.8cm 0.8cm 0.8cm},clip,height=13mm,width=13mm]{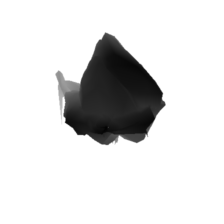}
        \includegraphics[trim={0.8cm 0.8cm 0.8cm 0.8cm},clip,height=13mm,width=13mm]{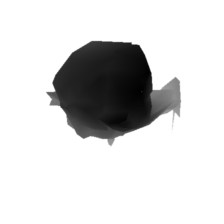} \\
        \raisebox{5.5mm}{\makebox[13mm][l]{\small{Proposed}}}
        \includegraphics[trim={0.8cm 0.8cm 0.8cm 0.8cm},clip,height=13mm,width=13mm]{./figure/images/9c7527d5d1fe9047f155d75bbf62b80_00.png}
        \includegraphics[trim={0.8cm 0.8cm 0.8cm 0.8cm},clip,height=13mm,width=13mm]{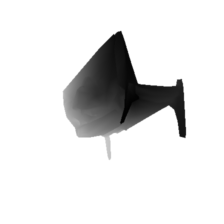}
        \includegraphics[trim={0.8cm 0.8cm 0.8cm 0.8cm},clip,height=13mm,width=13mm]{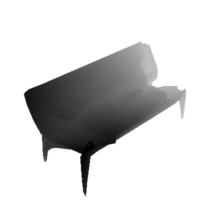}
        \includegraphics[trim={0.8cm 0.8cm 0.8cm 0.8cm},clip,height=13mm,width=13mm]{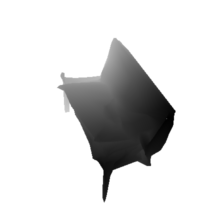}
        \includegraphics[trim={0.8cm 0.8cm 0.8cm 0.8cm},clip,height=13mm,width=13mm]{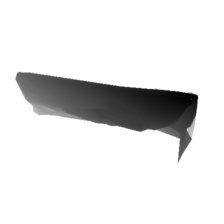} \\
        \raisebox{5.5mm}{\makebox[13mm][l]{\small{Baseline}}}
        \includegraphics[trim={0.8cm 0.8cm 0.8cm 0.8cm},clip,height=13mm,width=13mm]{./figure/images/9c7527d5d1fe9047f155d75bbf62b80_00.png}
        \includegraphics[trim={0.8cm 0.8cm 0.8cm 0.8cm},clip,height=13mm,width=13mm]{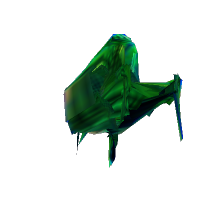}
        \includegraphics[trim={0.8cm 0.8cm 0.8cm 0.8cm},clip,height=13mm,width=13mm]{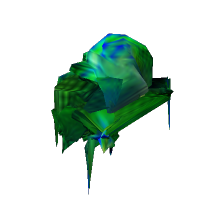}
        \includegraphics[trim={0.8cm 0.8cm 0.8cm 0.8cm},clip,height=13mm,width=13mm]{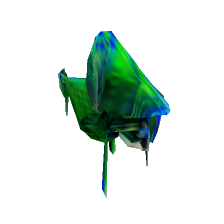}
        \includegraphics[trim={0.8cm 0.8cm 0.8cm 0.8cm},clip,height=13mm,width=13mm]{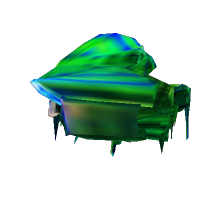} \\
        \raisebox{5.5mm}{\makebox[13mm][l]{\small{Proposed}}}
        \includegraphics[trim={0.8cm 0.8cm 0.8cm 0.8cm},clip,height=13mm,width=13mm]{./figure/images/9c7527d5d1fe9047f155d75bbf62b80_00.png}
        \includegraphics[trim={0.8cm 0.8cm 0.8cm 0.8cm},clip,height=13mm,width=13mm]{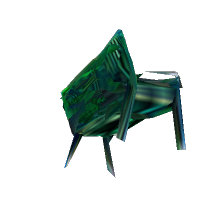}
        \includegraphics[trim={0.8cm 0.8cm 0.8cm 0.8cm},clip,height=13mm,width=13mm]{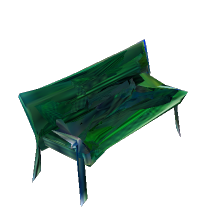}
        \includegraphics[trim={0.8cm 0.8cm 0.8cm 0.8cm},clip,height=13mm,width=13mm]{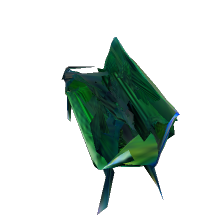}
        \includegraphics[trim={0.8cm 0.8cm 0.8cm 0.8cm},clip,height=13mm,width=13mm]{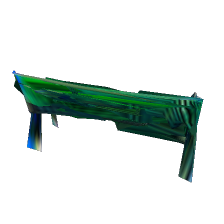} \\
        \makebox[13mm][c]{}
        \makebox[13mm][c]{\small{(a)}}
        \makebox[13mm][c]{\small{(b)}}
        \makebox[13mm][c]{\small{(c)}}
        \makebox[13mm][c]{\small{(d)}}
        \makebox[13mm][c]{\small{(e)}}
    \end{center}
    \caption{Examples on the ShapeNet dataset in single-view training. Panels (a--e) are the same as in Figure~\ref{fig:shapenet}. This figure corresponds to Section~\ref{sec:exp_single_view}. }
    \label{fig:shapenet_a1}
\end{figure}

\begin{figure}[!t]
    \begin{center}
        \raisebox{5.5mm}{\makebox[13mm][l]{\small{Baseline}}}
        \includegraphics[trim={0.8cm 0.8cm 0.8cm 0.8cm},clip,height=13mm,width=13mm]{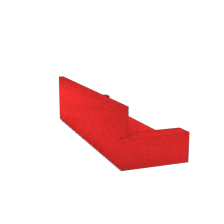}
        \includegraphics[trim={0.8cm 0.8cm 0.8cm 0.8cm},clip,height=13mm,width=13mm]{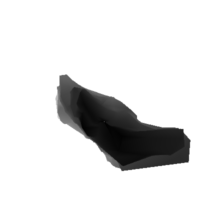}
        \includegraphics[trim={0.8cm 0.8cm 0.8cm 0.8cm},clip,height=13mm,width=13mm]{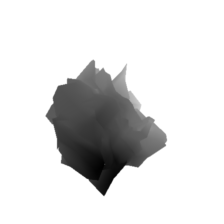}
        \includegraphics[trim={0.8cm 0.8cm 0.8cm 0.8cm},clip,height=13mm,width=13mm]{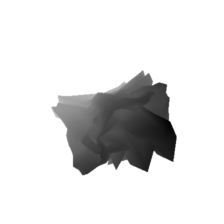}
        \includegraphics[trim={0.8cm 0.8cm 0.8cm 0.8cm},clip,height=13mm,width=13mm]{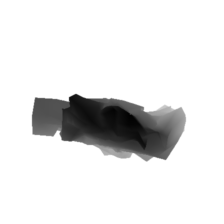} \\
        \raisebox{5.5mm}{\makebox[13mm][l]{\small{Proposed}}}
        \includegraphics[trim={0.8cm 0.8cm 0.8cm 0.8cm},clip,height=13mm,width=13mm]{./figure/images/20222a2bd14ea9609e489c1cf724666f_00.png}
        \includegraphics[trim={0.8cm 0.8cm 0.8cm 0.8cm},clip,height=13mm,width=13mm]{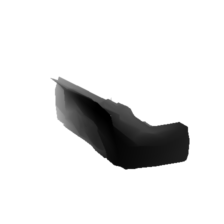}
        \includegraphics[trim={0.8cm 0.8cm 0.8cm 0.8cm},clip,height=13mm,width=13mm]{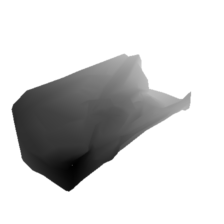}
        \includegraphics[trim={0.8cm 0.8cm 0.8cm 0.8cm},clip,height=13mm,width=13mm]{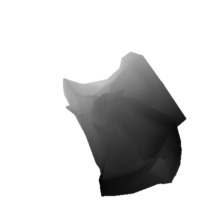}
        \includegraphics[trim={0.8cm 0.8cm 0.8cm 0.8cm},clip,height=13mm,width=13mm]{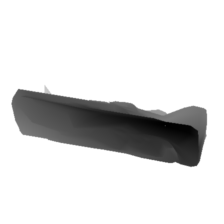} \\
        \raisebox{5.5mm}{\makebox[13mm][l]{\small{Baseline}}}
        \includegraphics[trim={0.8cm 0.8cm 0.8cm 0.8cm},clip,height=13mm,width=13mm]{./figure/images/20222a2bd14ea9609e489c1cf724666f_00.png}
        \includegraphics[trim={0.8cm 0.8cm 0.8cm 0.8cm},clip,height=13mm,width=13mm]{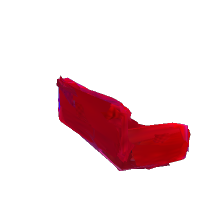}
        \includegraphics[trim={0.8cm 0.8cm 0.8cm 0.8cm},clip,height=13mm,width=13mm]{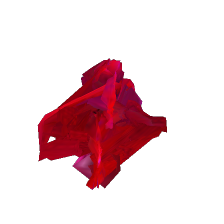}
        \includegraphics[trim={0.8cm 0.8cm 0.8cm 0.8cm},clip,height=13mm,width=13mm]{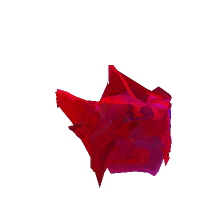}
        \includegraphics[trim={0.8cm 0.8cm 0.8cm 0.8cm},clip,height=13mm,width=13mm]{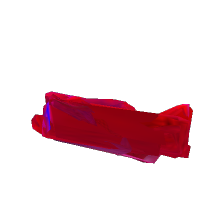} \\
        \raisebox{5.5mm}{\makebox[13mm][l]{\small{Proposed}}}
        \includegraphics[trim={0.8cm 0.8cm 0.8cm 0.8cm},clip,height=13mm,width=13mm]{./figure/images/20222a2bd14ea9609e489c1cf724666f_00.png}
        \includegraphics[trim={0.8cm 0.8cm 0.8cm 0.8cm},clip,height=13mm,width=13mm]{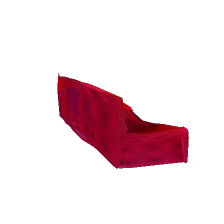}
        \includegraphics[trim={0.8cm 0.8cm 0.8cm 0.8cm},clip,height=13mm,width=13mm]{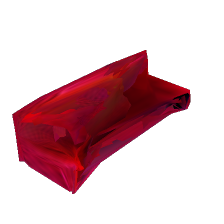}
        \includegraphics[trim={0.8cm 0.8cm 0.8cm 0.8cm},clip,height=13mm,width=13mm]{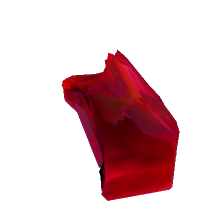}
        \includegraphics[trim={0.8cm 0.8cm 0.8cm 0.8cm},clip,height=13mm,width=13mm]{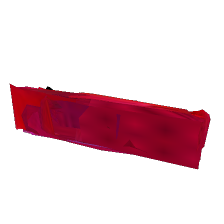} \\
        \raisebox{5.5mm}{\makebox[13mm][l]{\small{Baseline}}}
        \includegraphics[trim={0.8cm 0.8cm 0.8cm 0.8cm},clip,height=13mm,width=13mm]{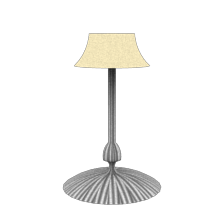}
        \includegraphics[trim={0.8cm 0.8cm 0.8cm 0.8cm},clip,height=13mm,width=13mm]{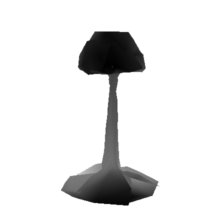}
        \includegraphics[trim={0.8cm 0.8cm 0.8cm 0.8cm},clip,height=13mm,width=13mm]{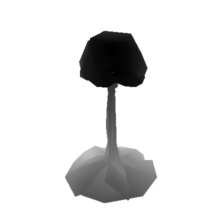}
        \includegraphics[trim={0.8cm 0.8cm 0.8cm 0.8cm},clip,height=13mm,width=13mm]{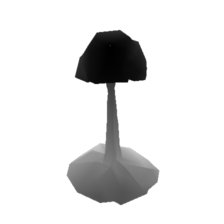}
        \includegraphics[trim={0.8cm 0.8cm 0.8cm 0.8cm},clip,height=13mm,width=13mm]{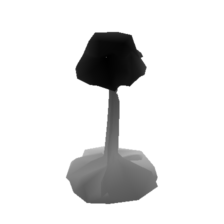} \\
        \raisebox{5.5mm}{\makebox[13mm][l]{\small{Proposed}}}
        \includegraphics[trim={0.8cm 0.8cm 0.8cm 0.8cm},clip,height=13mm,width=13mm]{./figure/images/4c266f2b866c59e761fef32872c6fa53_00.png}
        \includegraphics[trim={0.8cm 0.8cm 0.8cm 0.8cm},clip,height=13mm,width=13mm]{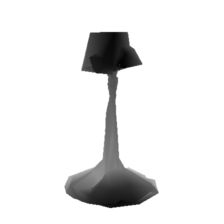}
        \includegraphics[trim={0.8cm 0.8cm 0.8cm 0.8cm},clip,height=13mm,width=13mm]{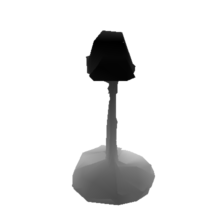}
        \includegraphics[trim={0.8cm 0.8cm 0.8cm 0.8cm},clip,height=13mm,width=13mm]{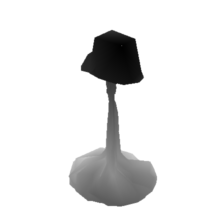}
        \includegraphics[trim={0.8cm 0.8cm 0.8cm 0.8cm},clip,height=13mm,width=13mm]{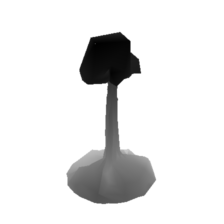} \\
        \raisebox{5.5mm}{\makebox[13mm][l]{\small{Baseline}}}
        \includegraphics[trim={0.8cm 0.8cm 0.8cm 0.8cm},clip,height=13mm,width=13mm]{./figure/images/4c266f2b866c59e761fef32872c6fa53_00.png}
        \includegraphics[trim={0.8cm 0.8cm 0.8cm 0.8cm},clip,height=13mm,width=13mm]{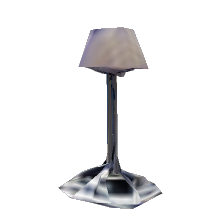}
        \includegraphics[trim={0.8cm 0.8cm 0.8cm 0.8cm},clip,height=13mm,width=13mm]{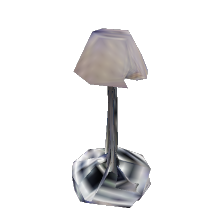}
        \includegraphics[trim={0.8cm 0.8cm 0.8cm 0.8cm},clip,height=13mm,width=13mm]{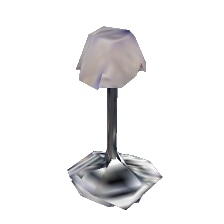}
        \includegraphics[trim={0.8cm 0.8cm 0.8cm 0.8cm},clip,height=13mm,width=13mm]{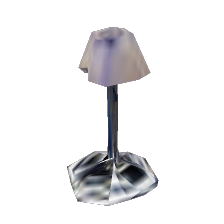} \\
        \raisebox{5.5mm}{\makebox[13mm][l]{\small{Proposed}}}
        \includegraphics[trim={0.8cm 0.8cm 0.8cm 0.8cm},clip,height=13mm,width=13mm]{./figure/images/4c266f2b866c59e761fef32872c6fa53_00.png}
        \includegraphics[trim={0.8cm 0.8cm 0.8cm 0.8cm},clip,height=13mm,width=13mm]{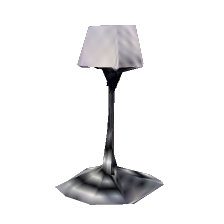}
        \includegraphics[trim={0.8cm 0.8cm 0.8cm 0.8cm},clip,height=13mm,width=13mm]{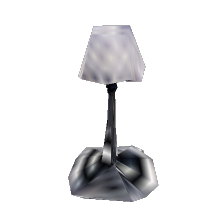}
        \includegraphics[trim={0.8cm 0.8cm 0.8cm 0.8cm},clip,height=13mm,width=13mm]{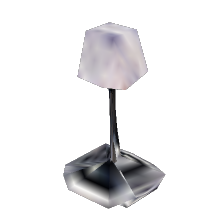}
        \includegraphics[trim={0.8cm 0.8cm 0.8cm 0.8cm},clip,height=13mm,width=13mm]{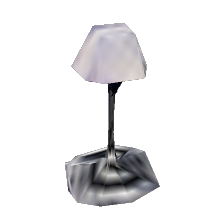} \\
        \raisebox{5.5mm}{\makebox[13mm][l]{\small{Baseline}}}
        \includegraphics[trim={0.8cm 0.8cm 0.8cm 0.8cm},clip,height=13mm,width=13mm]{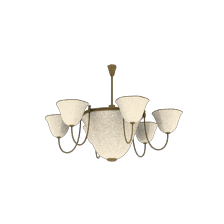}
        \includegraphics[trim={0.8cm 0.8cm 0.8cm 0.8cm},clip,height=13mm,width=13mm]{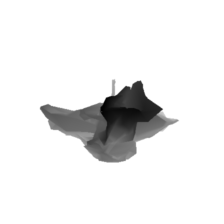}
        \includegraphics[trim={0.8cm 0.8cm 0.8cm 0.8cm},clip,height=13mm,width=13mm]{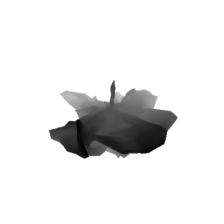}
        \includegraphics[trim={0.8cm 0.8cm 0.8cm 0.8cm},clip,height=13mm,width=13mm]{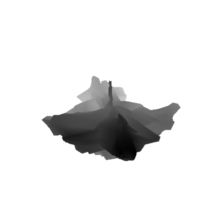}
        \includegraphics[trim={0.8cm 0.8cm 0.8cm 0.8cm},clip,height=13mm,width=13mm]{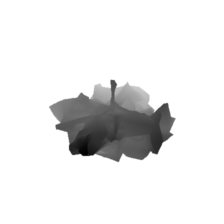} \\
        \raisebox{5.5mm}{\makebox[13mm][l]{\small{Proposed}}}
        \includegraphics[trim={0.8cm 0.8cm 0.8cm 0.8cm},clip,height=13mm,width=13mm]{./figure/images/f530508e27911aadabd4ed5db7667131_00.png}
        \includegraphics[trim={0.8cm 0.8cm 0.8cm 0.8cm},clip,height=13mm,width=13mm]{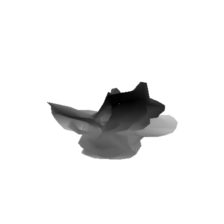}
        \includegraphics[trim={0.8cm 0.8cm 0.8cm 0.8cm},clip,height=13mm,width=13mm]{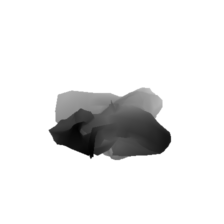}
        \includegraphics[trim={0.8cm 0.8cm 0.8cm 0.8cm},clip,height=13mm,width=13mm]{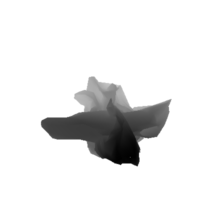}
        \includegraphics[trim={0.8cm 0.8cm 0.8cm 0.8cm},clip,height=13mm,width=13mm]{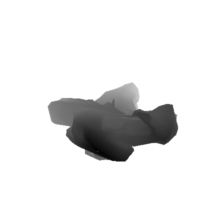} \\
        \raisebox{5.5mm}{\makebox[13mm][l]{\small{Baseline}}}
        \includegraphics[trim={0.8cm 0.8cm 0.8cm 0.8cm},clip,height=13mm,width=13mm]{./figure/images/f530508e27911aadabd4ed5db7667131_00.png}
        \includegraphics[trim={0.8cm 0.8cm 0.8cm 0.8cm},clip,height=13mm,width=13mm]{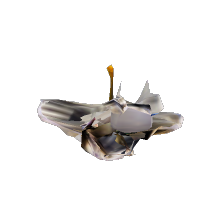}
        \includegraphics[trim={0.8cm 0.8cm 0.8cm 0.8cm},clip,height=13mm,width=13mm]{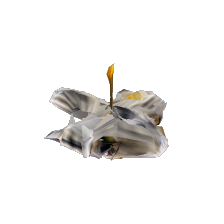}
        \includegraphics[trim={0.8cm 0.8cm 0.8cm 0.8cm},clip,height=13mm,width=13mm]{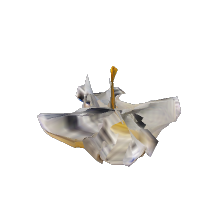}
        \includegraphics[trim={0.8cm 0.8cm 0.8cm 0.8cm},clip,height=13mm,width=13mm]{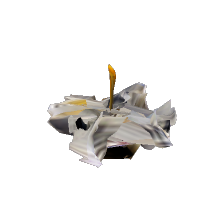} \\
        \raisebox{5.5mm}{\makebox[13mm][l]{\small{Proposed}}}
        \includegraphics[trim={0.8cm 0.8cm 0.8cm 0.8cm},clip,height=13mm,width=13mm]{./figure/images/f530508e27911aadabd4ed5db7667131_00.png}
        \includegraphics[trim={0.8cm 0.8cm 0.8cm 0.8cm},clip,height=13mm,width=13mm]{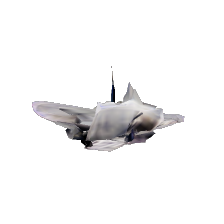}
        \includegraphics[trim={0.8cm 0.8cm 0.8cm 0.8cm},clip,height=13mm,width=13mm]{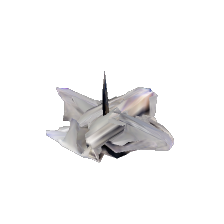}
        \includegraphics[trim={0.8cm 0.8cm 0.8cm 0.8cm},clip,height=13mm,width=13mm]{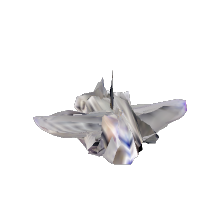}
        \includegraphics[trim={0.8cm 0.8cm 0.8cm 0.8cm},clip,height=13mm,width=13mm]{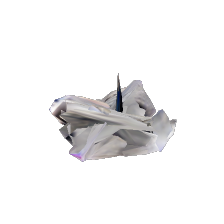} \\
        \makebox[13mm][c]{}
        \makebox[13mm][c]{\small{(a)}}
        \makebox[13mm][c]{\small{(b)}}
        \makebox[13mm][c]{\small{(c)}}
        \makebox[13mm][c]{\small{(d)}}
        \makebox[13mm][c]{\small{(e)}}
    \end{center}
    \caption{Examples on the ShapeNet dataset in single-view training. Notation of (a--e) is the same as in Figure~\ref{fig:shapenet}. This figure corresponds to Section~\ref{sec:exp_single_view}. }
    \label{fig:shapenet_a2}
\end{figure}

Figures~\ref{fig:shapenet_a1} and ~\ref{fig:shapenet_a2} show some reconstruction examples of {\it phone}, {\it display}, {\it bench}, {\it sofa}, and {\it lamp} categories on the ShapeNet dataset using single-view training. These figures correspond to the description in Section~\ref{sec:exp_single_view}. The difference among categories and the use of texture prediction can be examined from these figures.

%%%%%%%%%%%%%%%%%%%%%%%%%%%%%%%%%%%%%%%%%%%%%%%%%%%%%%%%%%%%%%%%%%%%%%%%%%%%%%%%
\subsection{Performance of each category by multi-view training}

\begin{table*}[!t]
    \begin{center}
        \small
        \begin{tabular}{ccc@{\hspace{0.6cm}}ccccccccccccc@{\hspace{0.6cm}}c}
            \toprule
            \STABBa{$N_v$} & \STABBa{VPL} &  \STABBa{TP} & \STABBa{airplane} & \STABBa{bench} & \STABBa{dresser} & \STABBa{car} & \STABBa{chair} & \STABBa{display} & \STABBa{lamp} & \STABBa{speaker} & \STABBa{rifle} & \STABBa{sofa} & \STABBa{table} & \STABBa{phone} & \STABBa{vessel} & \STABBa{all}  \\
            \hline
            $2$  &            &            & $.615$ & $.467$ & $.658$ & $.767$ & $.512$ & $.465$ & $.476$ & $.631$ & $.603$ & $.588$ & $.517$ & $.622$ & $.557$ & $.575$ \\
            $2$  & \checkmark &            & $.619$ & $.476$ & $.667$ & $.770$ & $.518$ & $.467$ & $.477$ & $.631$ & $.598$ & $.590$ & $.529$ & $.687$ & $.556$ & $.583$ \\
            $3$  &            &            & $.631$ & $.495$ & $.673$ & $.775$ & $.537$ & $.499$ & $.490$ & $.639$ & $.624$ & $.599$ & $.535$ & $.672$ & $.573$ & $.596$ \\
            $3$  & \checkmark &            & $.638$ & $.507$ & $.674$ & $.779$ & $.543$ & $.497$ & $.491$ & $.638$ & $.625$ & $.607$ & $.552$ & $.680$ & $.574$ & $.600$ \\
            $5$  &            &            & $.654$ & $.530$ & $.696$ & $.787$ & $.554$ & $.539$ & $.502$ & $.657$ & $.642$ & $.623$ & $.564$ & $.721$ & $.589$ & $.620$ \\
            $5$  & \checkmark &            & $.662$ & $.542$ & $.699$ & $.792$ & $.565$ & $.533$ & $.502$ & $.656$ & $.647$ & $.629$ & $.571$ & $.725$ & $.593$ & $.624$ \\
            $10$ &            &            & $.673$ & $.570$ & $.712$ & $.795$ & $.582$ & $.572$ & $.510$ & $.671$ & $.660$ & $.640$ & $.585$ & $.750$ & $.606$ & $.641$ \\
            $10$ & \checkmark &            & $.681$ & $.577$ & $.718$ & $.798$ & $.584$ & $.570$ & $.508$ & $.673$ & $.661$ & $.648$ & $.593$ & $.761$ & $.606$ & $.644$ \\
            $20$ &            &            & $.688$ & $.593$ & $.722$ & $.799$ & $.597$ & $.599$ & $.512$ & $.678$ & $.665$ & $.651$ & $.595$ & $.766$ & $.615$ & $.652$ \\
            $20$ & \checkmark &            & $.691$ & $.598$ & $.724$ & $.802$ & $.601$ & $.597$ & $.505$ & $.680$ & $.664$ & $.656$ & $.607$ & $.775$ & $.613$ & $.655$ \\
            \midrule
            $2$  &            & \checkmark & $.614$ & $.469$ & $.663$ & $.768$ & $.511$ & $.475$ & $.477$ & $.624$ & $.605$ & $.582$ & $.523$ & $.649$ & $.560$ & $.579$ \\
            $2$  & \checkmark & \checkmark & $.618$ & $.481$ & $.666$ & $.772$ & $.522$ & $.476$ & $.480$ & $.631$ & $.607$ & $.589$ & $.529$ & $.678$ & $.557$ & $.585$ \\
            $20$ &            & \checkmark & $.685$ & $.588$ & $.723$ & $.799$ & $.597$ & $.589$ & $.508$ & $.674$ & $.663$ & $.648$ & $.603$ & $.762$ & $.615$ & $.650$ \\
            $20$ & \checkmark & \checkmark & $.701$ & $.585$ & $.723$ & $.802$ & $.604$ & $.593$ & $.502$ & $.659$ & $.661$ & $.667$ & $.609$ & $.746$ & $.631$ & $.653$ \\
            \bottomrule
        \end{tabular}
    \end{center}
    \caption{IoU of multi-view training on the ShapeNet dataset dataset. This table corresponds Section~\ref{sec:exp_multi_view}. VPL: proposed view prior learning. TP: texture prediction. }
    \label{table:appendix_shapenet_multi_view_category}
\end{table*}

In Section~\ref{sec:exp_multi_view}, only the average performance in all categories on the ShapeNet dataset has been reported. Table~\ref{table:appendix_shapenet_multi_view_category} shows reconstruction accuracy of each category.

%%%%%%%%%%%%%%%%%%%%%%%%%%%%%%%%%%%%%%%%%%%%%%%%%%%%%%%%%%%%%%%%%%%%%%%%%%%%%%%%
\subsection{Additional examples of multi-view training}

\begin{figure}[!t]
    \begin{center}
        \includegraphics[trim={0.8cm 0.8cm 0.8cm 0.8cm},clip,height=13mm,width=13mm]{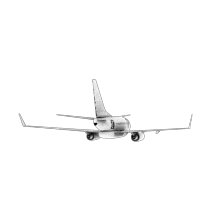}
        \includegraphics[trim={0.8cm 0.8cm 0.8cm 0.8cm},clip,height=13mm,width=13mm]{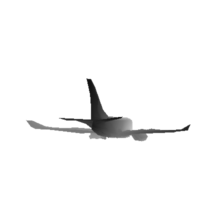}
        \includegraphics[trim={0.8cm 0.8cm 0.8cm 0.8cm},clip,height=13mm,width=13mm]{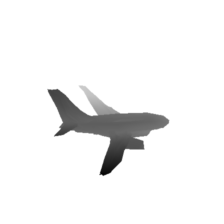}
        \includegraphics[trim={0.8cm 0.8cm 0.8cm 0.8cm},clip,height=13mm,width=13mm]{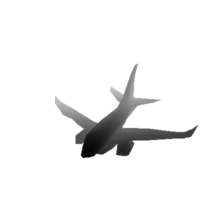}
        \includegraphics[trim={0.8cm 0.8cm 0.8cm 0.8cm},clip,height=13mm,width=13mm]{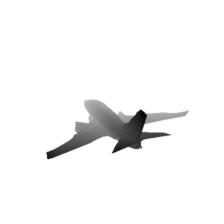} \\
        \includegraphics[trim={0.8cm 0.8cm 0.8cm 0.8cm},clip,height=13mm,width=13mm]{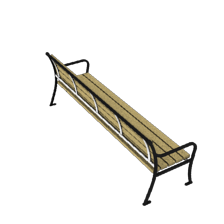}
        \includegraphics[trim={0.8cm 0.8cm 0.8cm 0.8cm},clip,height=13mm,width=13mm]{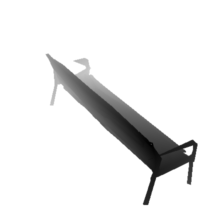}
        \includegraphics[trim={0.8cm 0.8cm 0.8cm 0.8cm},clip,height=13mm,width=13mm]{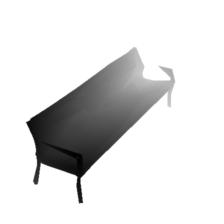}
        \includegraphics[trim={0.8cm 0.8cm 0.8cm 0.8cm},clip,height=13mm,width=13mm]{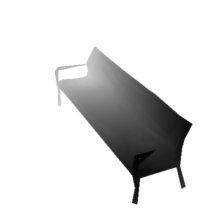}
        \includegraphics[trim={0.8cm 0.8cm 0.8cm 0.8cm},clip,height=13mm,width=13mm]{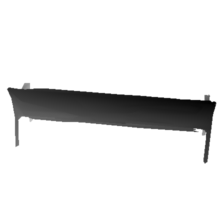} \\
        \includegraphics[trim={0.8cm 0.8cm 0.8cm 0.8cm},clip,height=13mm,width=13mm]{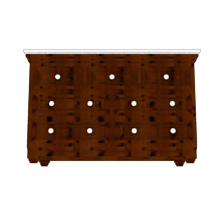}
        \includegraphics[trim={0.8cm 0.8cm 0.8cm 0.8cm},clip,height=13mm,width=13mm]{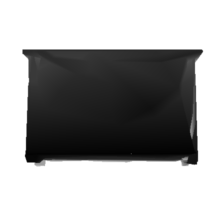}
        \includegraphics[trim={0.8cm 0.8cm 0.8cm 0.8cm},clip,height=13mm,width=13mm]{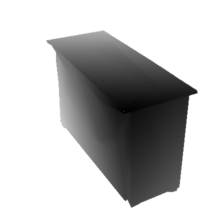}
        \includegraphics[trim={0.8cm 0.8cm 0.8cm 0.8cm},clip,height=13mm,width=13mm]{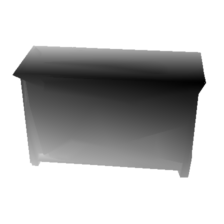}
        \includegraphics[trim={0.8cm 0.8cm 0.8cm 0.8cm},clip,height=13mm,width=13mm]{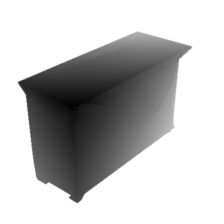} \\
        \includegraphics[trim={0.8cm 0.8cm 0.8cm 0.8cm},clip,height=13mm,width=13mm]{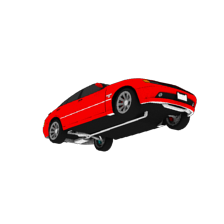}
        \includegraphics[trim={0.8cm 0.8cm 0.8cm 0.8cm},clip,height=13mm,width=13mm]{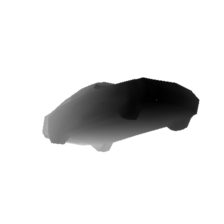}
        \includegraphics[trim={0.8cm 0.8cm 0.8cm 0.8cm},clip,height=13mm,width=13mm]{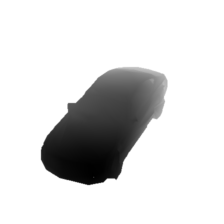}
        \includegraphics[trim={0.8cm 0.8cm 0.8cm 0.8cm},clip,height=13mm,width=13mm]{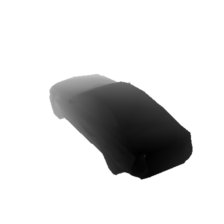}
        \includegraphics[trim={0.8cm 0.8cm 0.8cm 0.8cm},clip,height=13mm,width=13mm]{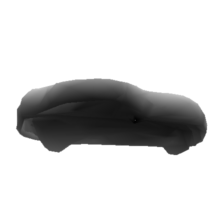} \\
        \includegraphics[trim={0.8cm 0.8cm 0.8cm 0.8cm},clip,height=13mm,width=13mm]{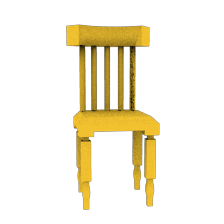}
        \includegraphics[trim={0.8cm 0.8cm 0.8cm 0.8cm},clip,height=13mm,width=13mm]{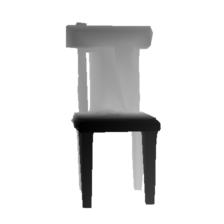}
        \includegraphics[trim={0.8cm 0.8cm 0.8cm 0.8cm},clip,height=13mm,width=13mm]{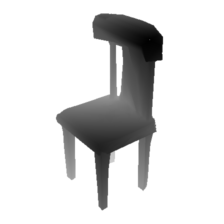}
        \includegraphics[trim={0.8cm 0.8cm 0.8cm 0.8cm},clip,height=13mm,width=13mm]{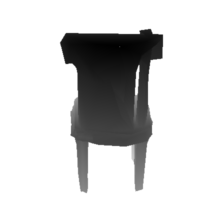}
        \includegraphics[trim={0.8cm 0.8cm 0.8cm 0.8cm},clip,height=13mm,width=13mm]{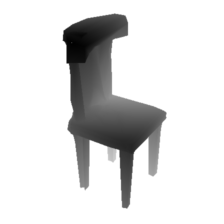} \\
        \includegraphics[trim={0.8cm 0.8cm 0.8cm 0.8cm},clip,height=13mm,width=13mm]{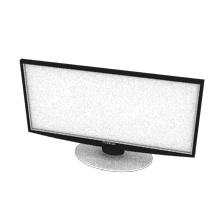}
        \includegraphics[trim={0.8cm 0.8cm 0.8cm 0.8cm},clip,height=13mm,width=13mm]{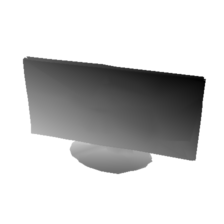}
        \includegraphics[trim={0.8cm 0.8cm 0.8cm 0.8cm},clip,height=13mm,width=13mm]{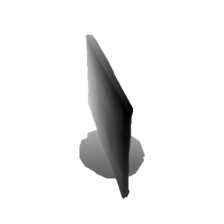}
        \includegraphics[trim={0.8cm 0.8cm 0.8cm 0.8cm},clip,height=13mm,width=13mm]{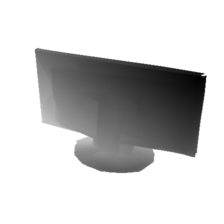}
        \includegraphics[trim={0.8cm 0.8cm 0.8cm 0.8cm},clip,height=13mm,width=13mm]{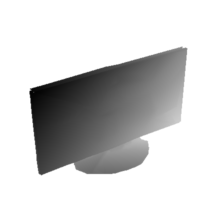} \\
        \includegraphics[trim={0.8cm 0.8cm 0.8cm 0.8cm},clip,height=13mm,width=13mm]{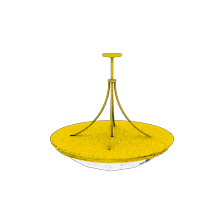}
        \includegraphics[trim={0.8cm 0.8cm 0.8cm 0.8cm},clip,height=13mm,width=13mm]{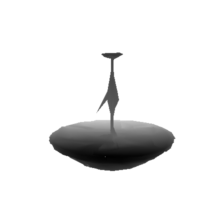}
        \includegraphics[trim={0.8cm 0.8cm 0.8cm 0.8cm},clip,height=13mm,width=13mm]{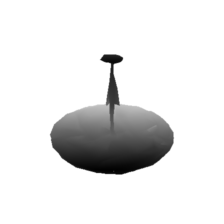}
        \includegraphics[trim={0.8cm 0.8cm 0.8cm 0.8cm},clip,height=13mm,width=13mm]{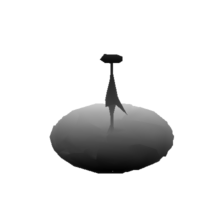}
        \includegraphics[trim={0.8cm 0.8cm 0.8cm 0.8cm},clip,height=13mm,width=13mm]{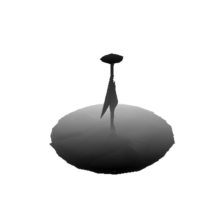} \\
        \includegraphics[trim={0.8cm 0.8cm 0.8cm 0.8cm},clip,height=13mm,width=13mm]{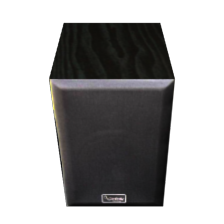}
        \includegraphics[trim={0.8cm 0.8cm 0.8cm 0.8cm},clip,height=13mm,width=13mm]{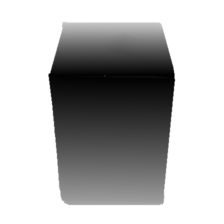}
        \includegraphics[trim={0.8cm 0.8cm 0.8cm 0.8cm},clip,height=13mm,width=13mm]{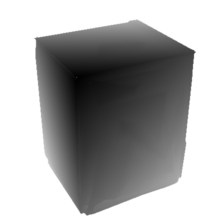}
        \includegraphics[trim={0.8cm 0.8cm 0.8cm 0.8cm},clip,height=13mm,width=13mm]{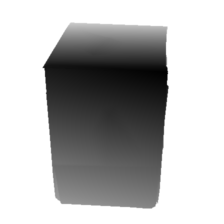}
        \includegraphics[trim={0.8cm 0.8cm 0.8cm 0.8cm},clip,height=13mm,width=13mm]{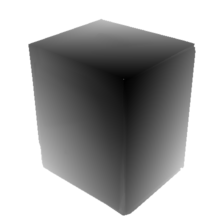} \\
        \includegraphics[trim={0.8cm 0.8cm 0.8cm 0.8cm},clip,height=13mm,width=13mm]{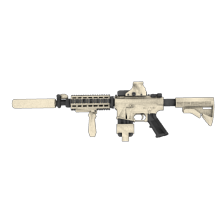}
        \includegraphics[trim={0.8cm 0.8cm 0.8cm 0.8cm},clip,height=13mm,width=13mm]{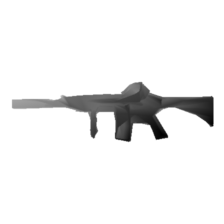}
        \includegraphics[trim={0.8cm 0.8cm 0.8cm 0.8cm},clip,height=13mm,width=13mm]{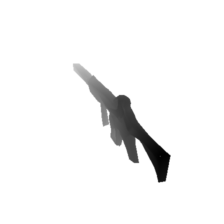}
        \includegraphics[trim={0.8cm 0.8cm 0.8cm 0.8cm},clip,height=13mm,width=13mm]{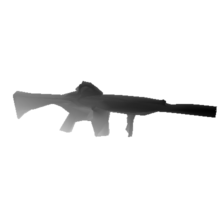}
        \includegraphics[trim={0.8cm 0.8cm 0.8cm 0.8cm},clip,height=13mm,width=13mm]{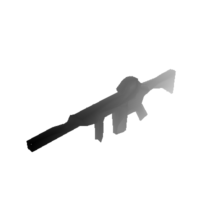} \\
        \includegraphics[trim={0.8cm 0.8cm 0.8cm 0.8cm},clip,height=13mm,width=13mm]{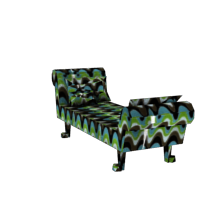}
        \includegraphics[trim={0.8cm 0.8cm 0.8cm 0.8cm},clip,height=13mm,width=13mm]{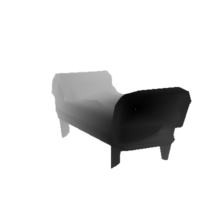}
        \includegraphics[trim={0.8cm 0.8cm 0.8cm 0.8cm},clip,height=13mm,width=13mm]{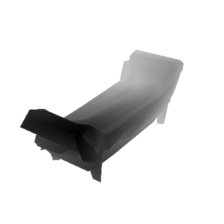}
        \includegraphics[trim={0.8cm 0.8cm 0.8cm 0.8cm},clip,height=13mm,width=13mm]{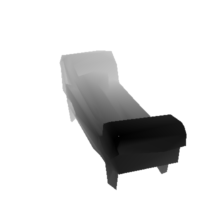}
        \includegraphics[trim={0.8cm 0.8cm 0.8cm 0.8cm},clip,height=13mm,width=13mm]{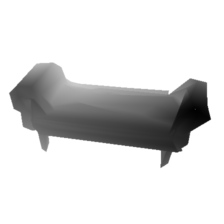} \\
        \includegraphics[trim={0.8cm 0.8cm 0.8cm 0.8cm},clip,height=13mm,width=13mm]{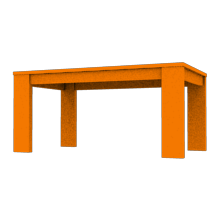}
        \includegraphics[trim={0.8cm 0.8cm 0.8cm 0.8cm},clip,height=13mm,width=13mm]{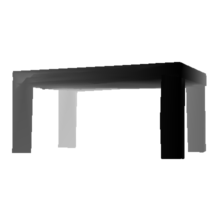}
        \includegraphics[trim={0.8cm 0.8cm 0.8cm 0.8cm},clip,height=13mm,width=13mm]{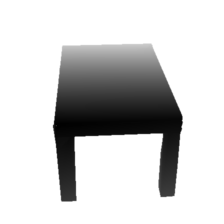}
        \includegraphics[trim={0.8cm 0.8cm 0.8cm 0.8cm},clip,height=13mm,width=13mm]{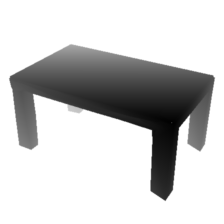}
        \includegraphics[trim={0.8cm 0.8cm 0.8cm 0.8cm},clip,height=13mm,width=13mm]{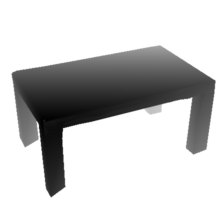} \\
        \includegraphics[trim={0.8cm 0.8cm 0.8cm 0.8cm},clip,height=13mm,width=13mm]{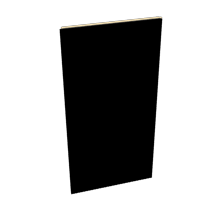}
        \includegraphics[trim={0.8cm 0.8cm 0.8cm 0.8cm},clip,height=13mm,width=13mm]{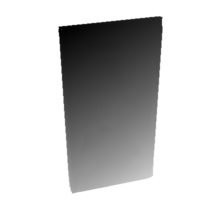}
        \includegraphics[trim={0.8cm 0.8cm 0.8cm 0.8cm},clip,height=13mm,width=13mm]{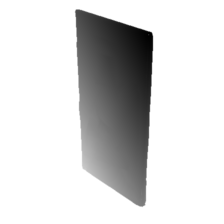}
        \includegraphics[trim={0.8cm 0.8cm 0.8cm 0.8cm},clip,height=13mm,width=13mm]{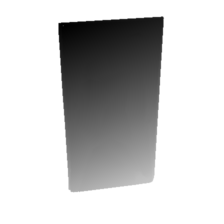}
        \includegraphics[trim={0.8cm 0.8cm 0.8cm 0.8cm},clip,height=13mm,width=13mm]{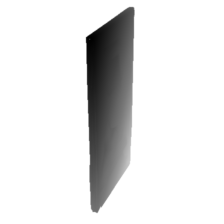} \\
        \includegraphics[trim={0.8cm 0.8cm 0.8cm 0.8cm},clip,height=13mm,width=13mm]{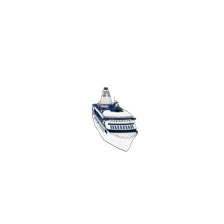}
        \includegraphics[trim={0.8cm 0.8cm 0.8cm 0.8cm},clip,height=13mm,width=13mm]{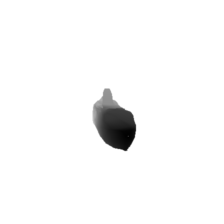}
        \includegraphics[trim={0.8cm 0.8cm 0.8cm 0.8cm},clip,height=13mm,width=13mm]{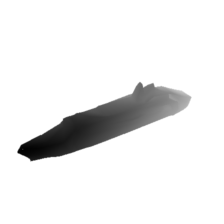}
        \includegraphics[trim={0.8cm 0.8cm 0.8cm 0.8cm},clip,height=13mm,width=13mm]{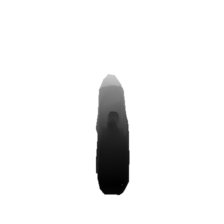}
        \includegraphics[trim={0.8cm 0.8cm 0.8cm 0.8cm},clip,height=13mm,width=13mm]{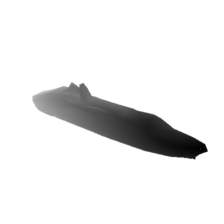} \\
        \makebox[13mm][c]{\small{(a)}}
        \makebox[13mm][c]{\small{(b)}}
        \makebox[13mm][c]{\small{(c)}}
        \makebox[13mm][c]{\small{(d)}}
        \makebox[13mm][c]{\small{(e)}}
    \end{center}
    \caption{Examples of thirteen categories on the ShapeNet dataset by multi-view training ($N_v$ = 20) without texture prediction. Panels (a--e) are the same as in Figure~\ref{fig:shapenet}.}
    \label{fig:appendix_shapenet_nv20_silhouette}
\end{figure}

\begin{figure}[!t]
    \begin{center}
        \includegraphics[trim={0.8cm 0.8cm 0.8cm 0.8cm},clip,height=13mm,width=13mm]{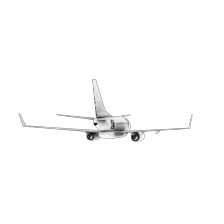}
        \includegraphics[trim={0.8cm 0.8cm 0.8cm 0.8cm},clip,height=13mm,width=13mm]{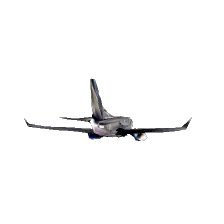}
        \includegraphics[trim={0.8cm 0.8cm 0.8cm 0.8cm},clip,height=13mm,width=13mm]{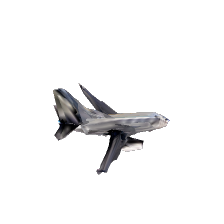}
        \includegraphics[trim={0.8cm 0.8cm 0.8cm 0.8cm},clip,height=13mm,width=13mm]{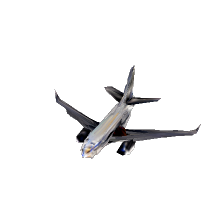}
        \includegraphics[trim={0.8cm 0.8cm 0.8cm 0.8cm},clip,height=13mm,width=13mm]{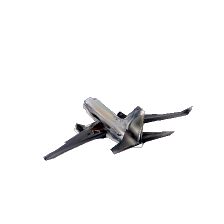} \\
        \includegraphics[trim={0.8cm 0.8cm 0.8cm 0.8cm},clip,height=13mm,width=13mm]{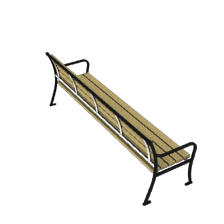}
        \includegraphics[trim={0.8cm 0.8cm 0.8cm 0.8cm},clip,height=13mm,width=13mm]{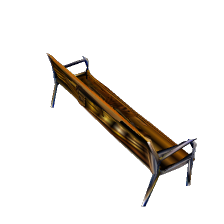}
        \includegraphics[trim={0.8cm 0.8cm 0.8cm 0.8cm},clip,height=13mm,width=13mm]{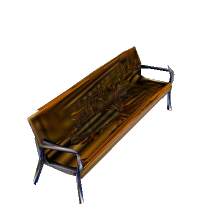}
        \includegraphics[trim={0.8cm 0.8cm 0.8cm 0.8cm},clip,height=13mm,width=13mm]{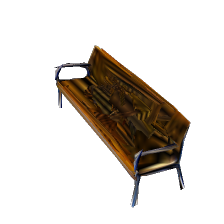}
        \includegraphics[trim={0.8cm 0.8cm 0.8cm 0.8cm},clip,height=13mm,width=13mm]{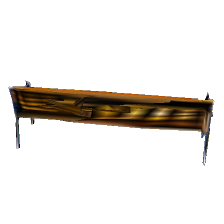} \\
        \includegraphics[trim={0.8cm 0.8cm 0.8cm 0.8cm},clip,height=13mm,width=13mm]{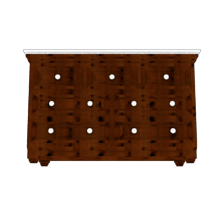}
        \includegraphics[trim={0.8cm 0.8cm 0.8cm 0.8cm},clip,height=13mm,width=13mm]{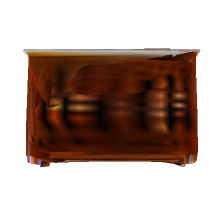}
        \includegraphics[trim={0.8cm 0.8cm 0.8cm 0.8cm},clip,height=13mm,width=13mm]{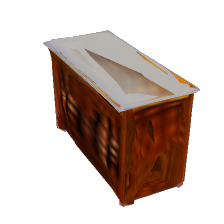}
        \includegraphics[trim={0.8cm 0.8cm 0.8cm 0.8cm},clip,height=13mm,width=13mm]{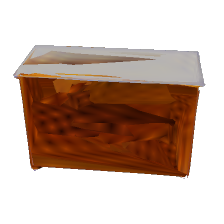}
        \includegraphics[trim={0.8cm 0.8cm 0.8cm 0.8cm},clip,height=13mm,width=13mm]{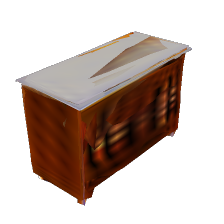} \\
        \includegraphics[trim={0.8cm 0.8cm 0.8cm 0.8cm},clip,height=13mm,width=13mm]{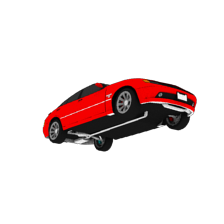}
        \includegraphics[trim={0.8cm 0.8cm 0.8cm 0.8cm},clip,height=13mm,width=13mm]{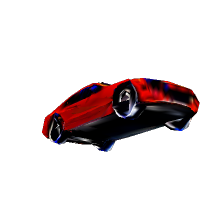}
        \includegraphics[trim={0.8cm 0.8cm 0.8cm 0.8cm},clip,height=13mm,width=13mm]{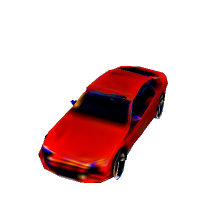}
        \includegraphics[trim={0.8cm 0.8cm 0.8cm 0.8cm},clip,height=13mm,width=13mm]{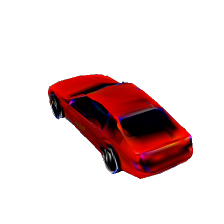}
        \includegraphics[trim={0.8cm 0.8cm 0.8cm 0.8cm},clip,height=13mm,width=13mm]{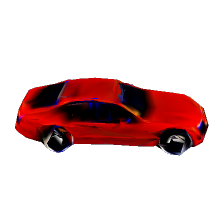} \\
        \includegraphics[trim={0.8cm 0.8cm 0.8cm 0.8cm},clip,height=13mm,width=13mm]{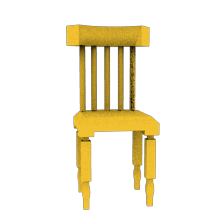}
        \includegraphics[trim={0.8cm 0.8cm 0.8cm 0.8cm},clip,height=13mm,width=13mm]{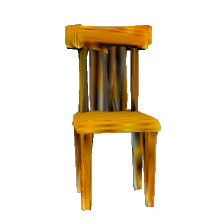}
        \includegraphics[trim={0.8cm 0.8cm 0.8cm 0.8cm},clip,height=13mm,width=13mm]{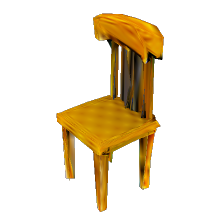}
        \includegraphics[trim={0.8cm 0.8cm 0.8cm 0.8cm},clip,height=13mm,width=13mm]{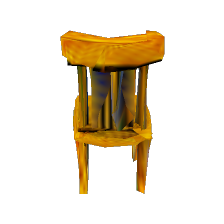}
        \includegraphics[trim={0.8cm 0.8cm 0.8cm 0.8cm},clip,height=13mm,width=13mm]{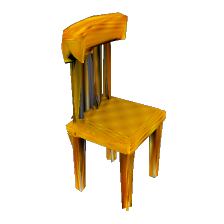} \\
        \includegraphics[trim={0.8cm 0.8cm 0.8cm 0.8cm},clip,height=13mm,width=13mm]{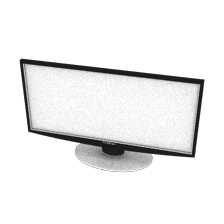}
        \includegraphics[trim={0.8cm 0.8cm 0.8cm 0.8cm},clip,height=13mm,width=13mm]{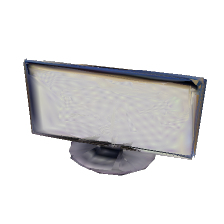}
        \includegraphics[trim={0.8cm 0.8cm 0.8cm 0.8cm},clip,height=13mm,width=13mm]{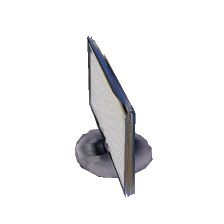}
        \includegraphics[trim={0.8cm 0.8cm 0.8cm 0.8cm},clip,height=13mm,width=13mm]{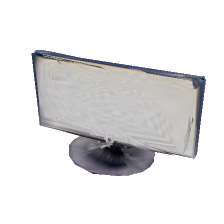}
        \includegraphics[trim={0.8cm 0.8cm 0.8cm 0.8cm},clip,height=13mm,width=13mm]{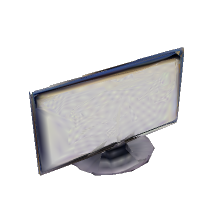} \\
        \includegraphics[trim={0.8cm 0.8cm 0.8cm 0.8cm},clip,height=13mm,width=13mm]{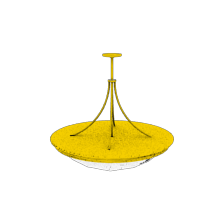}
        \includegraphics[trim={0.8cm 0.8cm 0.8cm 0.8cm},clip,height=13mm,width=13mm]{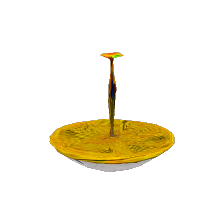}
        \includegraphics[trim={0.8cm 0.8cm 0.8cm 0.8cm},clip,height=13mm,width=13mm]{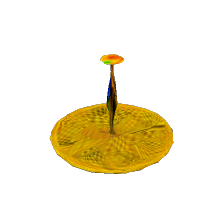}
        \includegraphics[trim={0.8cm 0.8cm 0.8cm 0.8cm},clip,height=13mm,width=13mm]{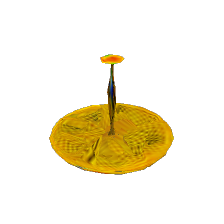}
        \includegraphics[trim={0.8cm 0.8cm 0.8cm 0.8cm},clip,height=13mm,width=13mm]{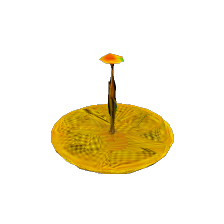} \\
        \includegraphics[trim={0.8cm 0.8cm 0.8cm 0.8cm},clip,height=13mm,width=13mm]{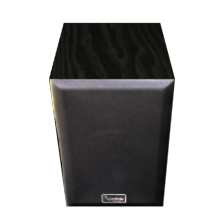}
        \includegraphics[trim={0.8cm 0.8cm 0.8cm 0.8cm},clip,height=13mm,width=13mm]{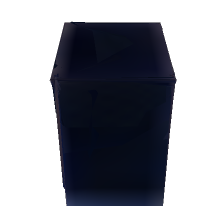}
        \includegraphics[trim={0.8cm 0.8cm 0.8cm 0.8cm},clip,height=13mm,width=13mm]{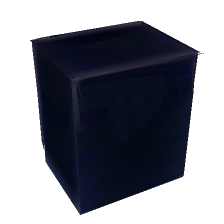}
        \includegraphics[trim={0.8cm 0.8cm 0.8cm 0.8cm},clip,height=13mm,width=13mm]{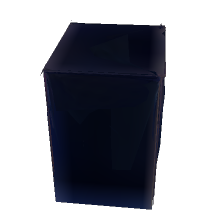}
        \includegraphics[trim={0.8cm 0.8cm 0.8cm 0.8cm},clip,height=13mm,width=13mm]{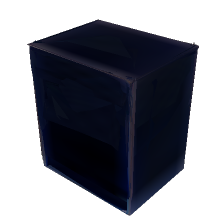} \\
        \includegraphics[trim={0.8cm 0.8cm 0.8cm 0.8cm},clip,height=13mm,width=13mm]{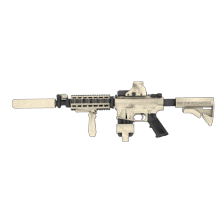}
        \includegraphics[trim={0.8cm 0.8cm 0.8cm 0.8cm},clip,height=13mm,width=13mm]{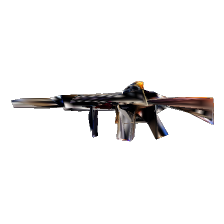}
        \includegraphics[trim={0.8cm 0.8cm 0.8cm 0.8cm},clip,height=13mm,width=13mm]{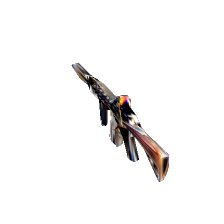}
        \includegraphics[trim={0.8cm 0.8cm 0.8cm 0.8cm},clip,height=13mm,width=13mm]{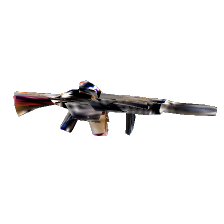}
        \includegraphics[trim={0.8cm 0.8cm 0.8cm 0.8cm},clip,height=13mm,width=13mm]{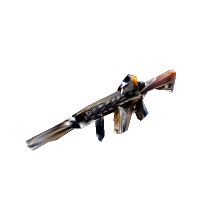} \\
        \includegraphics[trim={0.8cm 0.8cm 0.8cm 0.8cm},clip,height=13mm,width=13mm]{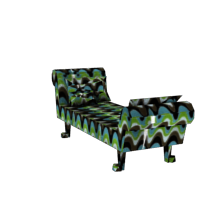}
        \includegraphics[trim={0.8cm 0.8cm 0.8cm 0.8cm},clip,height=13mm,width=13mm]{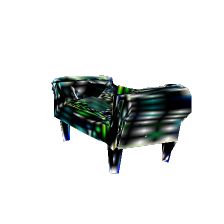}
        \includegraphics[trim={0.8cm 0.8cm 0.8cm 0.8cm},clip,height=13mm,width=13mm]{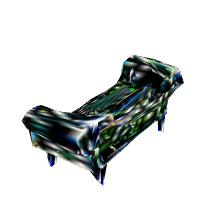}
        \includegraphics[trim={0.8cm 0.8cm 0.8cm 0.8cm},clip,height=13mm,width=13mm]{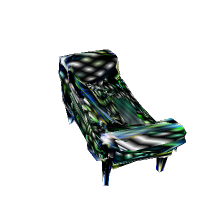}
        \includegraphics[trim={0.8cm 0.8cm 0.8cm 0.8cm},clip,height=13mm,width=13mm]{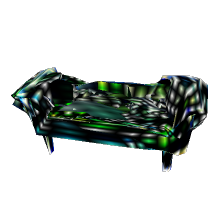} \\
        \includegraphics[trim={0.8cm 0.8cm 0.8cm 0.8cm},clip,height=13mm,width=13mm]{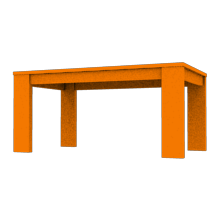}
        \includegraphics[trim={0.8cm 0.8cm 0.8cm 0.8cm},clip,height=13mm,width=13mm]{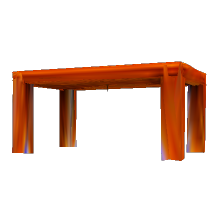}
        \includegraphics[trim={0.8cm 0.8cm 0.8cm 0.8cm},clip,height=13mm,width=13mm]{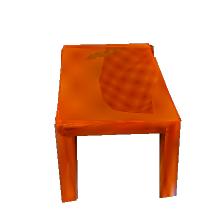}
        \includegraphics[trim={0.8cm 0.8cm 0.8cm 0.8cm},clip,height=13mm,width=13mm]{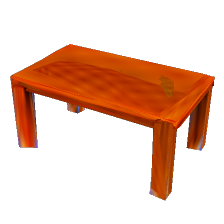}
        \includegraphics[trim={0.8cm 0.8cm 0.8cm 0.8cm},clip,height=13mm,width=13mm]{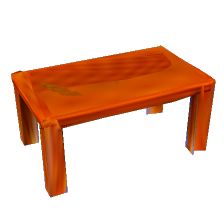} \\
        \includegraphics[trim={0.8cm 0.8cm 0.8cm 0.8cm},clip,height=13mm,width=13mm]{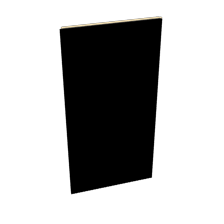}
        \includegraphics[trim={0.8cm 0.8cm 0.8cm 0.8cm},clip,height=13mm,width=13mm]{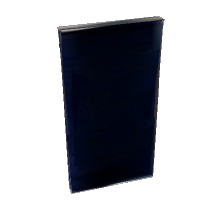}
        \includegraphics[trim={0.8cm 0.8cm 0.8cm 0.8cm},clip,height=13mm,width=13mm]{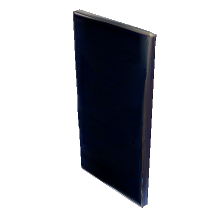}
        \includegraphics[trim={0.8cm 0.8cm 0.8cm 0.8cm},clip,height=13mm,width=13mm]{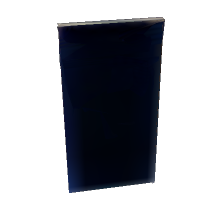}
        \includegraphics[trim={0.8cm 0.8cm 0.8cm 0.8cm},clip,height=13mm,width=13mm]{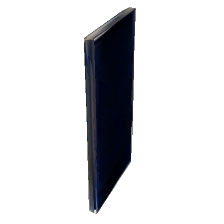} \\
        \includegraphics[trim={0.8cm 0.8cm 0.8cm 0.8cm},clip,height=13mm,width=13mm]{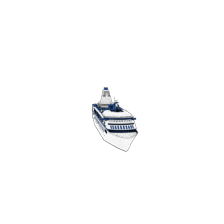}
        \includegraphics[trim={0.8cm 0.8cm 0.8cm 0.8cm},clip,height=13mm,width=13mm]{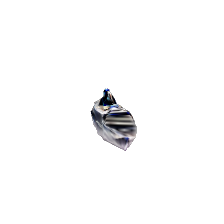}
        \includegraphics[trim={0.8cm 0.8cm 0.8cm 0.8cm},clip,height=13mm,width=13mm]{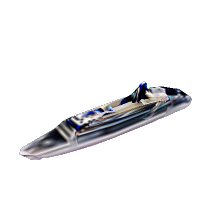}
        \includegraphics[trim={0.8cm 0.8cm 0.8cm 0.8cm},clip,height=13mm,width=13mm]{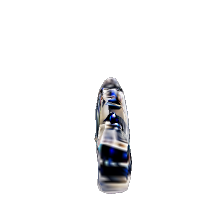}
        \includegraphics[trim={0.8cm 0.8cm 0.8cm 0.8cm},clip,height=13mm,width=13mm]{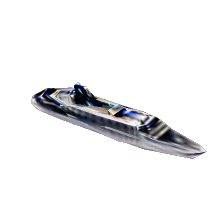} \\
        \makebox[13mm][c]{\small{(a)}}
        \makebox[13mm][c]{\small{(b)}}
        \makebox[13mm][c]{\small{(c)}}
        \makebox[13mm][c]{\small{(d)}}
        \makebox[13mm][c]{\small{(e)}}
    \end{center}
    \caption{Examples of thirteen categories on the ShapeNet dataset by multi-view training ($N_v$ = 20) with texture prediction. Panels (a--e) are the same as in Figure~\ref{fig:shapenet}.}
    \label{fig:appendix_shapenet_nv20_color}
\end{figure}

Figures~\ref{fig:appendix_shapenet_nv20_silhouette} and ~\ref{fig:appendix_shapenet_nv20_color} show results from our best performing models using multi-view training ($N_v = 20$) for those interested in the state-of-the-art performance on the ShapeNet dataset. As can be seen in the figure, high-quality 3D models with textures can be reconstructed without using 3D models for training. The mean IoU of our method without texture prediction is $65.5$, and the mean IoU of our method with texture prediction is $65.0$. In contrast to single-view training, texture prediction does not improve the performance in multi-view training for large $N_v$.

%%%%%%%%%%%%%%%%%%%%%%%%%%%%%%%%%%%%%%%%%%%%%%%%%%%%%%%%%%%%%%%%%%%%%%%%%%%%%%%%
\subsection{Evaluation using CD and EMD}

\begin{table*}[!t]
    \begin{center}
        \small
        \begin{tabular}{ccc@{\hspace{0.5cm}}ccccccccccccc@{\hspace{0.5cm}}c}
            \toprule
            \STABBa{VPL} & \STABBa{CC} &  \STABBa{TP} & \STABBa{airplane} & \STABBa{bench} & \STABBa{dresser} & \STABBa{car} & \STABBa{chair} & \STABBa{display} & \STABBa{lamp} & \STABBa{speaker} & \STABBa{rifle} & \STABBa{sofa} & \STABBa{table} & \STABBa{phone} & \STABBa{vessel} & \STABBa{all}  \\
            \hline
                       &            &            & $2.07$ & $9.69$ & $9.07$ & $4.91$ & $5.01$ & $8.72$ & $5.94$ & $11.49$ & $2.80$ & $9.05$ & $10.14$ & $4.36$ & $4.41$ & $6.74$ \\
            \checkmark &            &            & $2.33$ & $6.69$ & $7.73$ & $3.40$ & $4.94$ & $8.51$ & $7.81$ &  $9.24$ & $2.48$ & $7.28$ &  $5.88$ & $5.48$ & $3.76$ & $5.81$ \\
            \checkmark & \checkmark &            & $2.06$ & $5.45$ & $6.58$ & $2.83$ & $4.24$ & $\mathbf{5.11}$ & $6.31$ &  $9.57$ & $2.02$ & $5.65$ &  $6.56$ & $2.99$ & $3.17$ & $4.81$ \\
                       &            & \checkmark & $1.94$ & $8.17$ & $6.42$ & $4.17$ & $3.96$ & $6.78$ & $\mathbf{5.24}$ &  $7.34$ & $2.59$ & $7.65$ &  $6.21$ & $4.28$ & $3.95$ & $5.29$ \\
            \checkmark &            & \checkmark & $1.58$ & $3.97$ & $\mathbf{4.66}$ & $2.92$ & $\mathbf{3.24}$ & $5.48$ & $5.36$ &  $\mathbf{6.15}$ & $1.82$ & $5.19$ &  $\mathbf{3.55}$ & $\mathbf{2.59}$ & $3.23$ & $3.83$ \\
            \checkmark & \checkmark & \checkmark & $\mathbf{1.52}$ & $\mathbf{3.94}$ & $5.19$ & $\mathbf{2.76}$ & $3.42$ & $5.15$ & $5.37$ &  $6.38$ & $\mathbf{1.50}$ & $\mathbf{4.87}$ &  $3.64$ & $2.61$ & $\mathbf{2.85}$ & $\mathbf{3.78}$ \\
            \bottomrule
        \end{tabular}
    \end{center}
    \caption{Evaluation using CD. Points are uniformly sampled from surfaces. This table corresponds to Table~\ref{table:shapenet}. }
    \label{table:shapenet_cd_s}
\end{table*}

\begin{table*}[!t]
    \begin{center}
        \small
        \begin{tabular}{ccc@{\hspace{0.5cm}}ccccccccccccc@{\hspace{0.5cm}}c}
            \toprule
            \STABBa{VPL} & \STABBa{CC} &  \STABBa{TP} & \STABBa{airplane} & \STABBa{bench} & \STABBa{dresser} & \STABBa{car} & \STABBa{chair} & \STABBa{display} & \STABBa{lamp} & \STABBa{speaker} & \STABBa{rifle} & \STABBa{sofa} & \STABBa{table} & \STABBa{phone} & \STABBa{vessel} & \STABBa{all}  \\
            \hline
                       &            &            & $2.39$ & $9.52$ & $4.61$ & $2.66$ & $5.17$ & $12.29$ & $4.74$ & $3.86$ & $3.96$ & $8.32$ & $5.30$ & $5.93$ & $3.92$ & $5.59$ \\
            \checkmark &            &            & $2.56$ & $5.93$ & $2.04$ & $1.05$ & $3.88$ &  $6.77$ & $6.47$ & $3.20$ & $3.02$ & $6.29$ &  $3.77$ & $3.39$ & $2.79$ & $3.93$ \\
            \checkmark & \checkmark &            & $2.24$ & $4.34$ & $1.82$ & $0.76$ & $3.07$ &  $4.37$ & $5.58$ & $\mathbf{2.51}$ & $2.94$ & $4.38$ &  $3.49$ & $2.08$ & $2.85$ & $3.11$ \\
                       &            & \checkmark & $2.21$ & $8.04$ & $2.51$ & $2.66$ & $4.81$ & $5.35$ & $\mathbf{4.58}$ &  $3.23$ & $3.54$ & $7.70$ &  $4.67$ & $2.03$ & $3.53$ & $4.22$ \\
            \checkmark &            & \checkmark & $1.74$ & $3.74$ & $1.84$ & $0.72$ & $2.47$ & $\mathbf{4.02}$ & $4.69$ &  $2.72$ & $2.60$ & $4.20$ &  $2.97$ & $\mathbf{1.50}$ & $2.52$ & $2.75$ \\
            \checkmark & \checkmark & \checkmark & $\mathbf{1.66}$ & $\mathbf{3.71}$ & $\mathbf{1.72}$ & $\mathbf{0.71}$ & $\mathbf{2.40}$ & $4.04$ & $4.59$ &  $2.68$ & $\mathbf{2.20}$ & $\mathbf{3.95}$ &  $\mathbf{2.89}$ & $1.64$ & $\mathbf{2.31}$ & $\mathbf{2.65}$ \\
            \bottomrule
        \end{tabular}
    \end{center}
    \caption{Evaluation using CD. Points are uniformly sampled from volumes. This table corresponds to Table~\ref{table:shapenet}. }
    \label{table:shapenet_cd_v}
\end{table*}

\begin{table*}[!t]
    \begin{center}
        \small
        \begin{tabular}{ccc@{\hspace{0.5cm}}ccccccccccccc@{\hspace{0.5cm}}c}
            \toprule
            \STABBa{VPL} & \STABBa{CC} &  \STABBa{TP} & \STABBa{airplane} & \STABBa{bench} & \STABBa{dresser} & \STABBa{car} & \STABBa{chair} & \STABBa{display} & \STABBa{lamp} & \STABBa{speaker} & \STABBa{rifle} & \STABBa{sofa} & \STABBa{table} & \STABBa{phone} & \STABBa{vessel} & \STABBa{all}  \\
            \hline
                       &            &            & $14.8$ & $23.2$ & $\mathbf{24.9}$ & $\mathbf{21.1}$ & $\mathbf{21.5}$ & $\mathbf{22.2}$ & $23.2$ & $\mathbf{24.9}$ & $17.4$ & $24.0$ & $\mathbf{23.9}$ & $22.8$ & $19.2$ & $\mathbf{21.8}$ \\
            \checkmark &            &            & $14.8$ & $23.6$ & $29.3$ & $22.4$ & $22.3$ & $25.0$ & $24.4$ & $29.2$ & $\mathbf{16.9}$ & $26.1$ & $24.7$ & $27.1$ & $19.4$ & $23.5$ \\
            \checkmark & \checkmark &            & $14.5$ & $22.8$ & $25.2$ & $21.6$ & $22.9$ & $22.6$ & $\mathbf{23.1}$ & $25.2$ & $17.0$ & $\mathbf{23.9}$ & $24.8$ & $\mathbf{20.9}$ & $\mathbf{18.3}$ & $\mathbf{21.8}$ \\
                       &            & \checkmark & $14.7$ & $23.1$ & $27.0$ & $26.0$ & $23.3$ & $22.3$ & $24.0$ & $28.2$ & $17.3$ & $24.2$ & $25.5$ & $26.9$ & $19.7$ & $23.3$ \\
            \checkmark &            & \checkmark & $14.3$ & $25.7$ & $31.6$ & $23.0$ & $24.3$ & $32.8$ & $24.1$ & $31.0$ & $17.1$ & $26.8$ & $25.2$ & $28.7$ & $19.1$ & $24.9$ \\
            \checkmark & \checkmark & \checkmark & $\mathbf{14.2}$ & $\mathbf{22.5}$ & $30.7$ & $24.3$ & $23.0$ & $30.2$ & $24.1$ & $30.2$ & $17.0$ & $24.3$ & $24.9$ & $31.5$ & $19.0$ & $24.3$ \\
            \bottomrule
        \end{tabular}
    \end{center}
    \caption{Evaluation using EMD. Points are uniformly sampled from surfaces. This table corresponds to Table~\ref{table:shapenet}. }
    \label{table:shapenet_emd_s}
\end{table*}

\begin{table*}[!t]
    \begin{center}
        \small
        \begin{tabular}{ccc@{\hspace{0.5cm}}ccccccccccccc@{\hspace{0.5cm}}c}
            \toprule
            \STABBa{VPL} & \STABBa{CC} &  \STABBa{TP} & \STABBa{airplane} & \STABBa{bench} & \STABBa{dresser} & \STABBa{car} & \STABBa{chair} & \STABBa{display} & \STABBa{lamp} & \STABBa{speaker} & \STABBa{rifle} & \STABBa{sofa} & \STABBa{table} & \STABBa{phone} & \STABBa{vessel} & \STABBa{all}  \\
            \hline
                       &            &            & $0.21$ & $1.61$ & $2.17$ & $0.93$ & $1.27$ & $3.65$ & $0.58$ & $1.67$ & $0.27$ & $2.08$ & $1.52$ & $2.23$ & $0.61$ & $1.45$ \\
            \checkmark &            &            & $0.20$ & $0.92$ & $1.07$ & $0.44$ & $0.94$ & $1.12$ & $\mathbf{0.53}$ & $1.32$ & $0.20$ & $1.25$ & $1.00$ & $\mathbf{0.45}$ & $0.48$ & $0.76$ \\
            \checkmark & \checkmark &            & $0.19$ & $0.79$ & $1.12$ & $\mathbf{0.37}$ & $0.85$ & $1.49$ & $0.87$ & $\mathbf{1.17}$ & $0.22$ & $1.28$ & $1.13$ & $0.96$ & $0.48$ & $0.84$ \\
                       &            & \checkmark & $0.20$ & $1.17$ & $1.31$ & $0.92$ & $1.20$ & $1.36$ & $0.63$ & $1.57$ & $0.26$ & $1.90$ & $1.30$ & $0.71$ & $0.57$ & $1.01$ \\
            \checkmark &            & \checkmark & $\mathbf{0.17}$ & $\mathbf{0.66}$ & $1.08$ & $0.42$ & $0.82$ & $1.01$ & $0.65$ & $1.35$ & $0.19$ & $1.25$ & $\mathbf{0.84}$ & $0.47$ & $0.48$ & $0.72$ \\
            \checkmark & \checkmark & \checkmark & $\mathbf{0.17}$ & $\mathbf{0.66}$ & $\mathbf{1.06}$ & $0.41$ & $\mathbf{0.78}$ & $\mathbf{1.00}$ & $0.62$ & $1.35$ & $\mathbf{0.18}$ & $\mathbf{1.08}$ & $0.85$ & $0.50$ & $\mathbf{0.45}$ & $\mathbf{0.70}$ \\
            \bottomrule
        \end{tabular}
    \end{center}
    \caption{Evaluation using EMD. Points are uniformly sampled from volumes. This table corresponds to Table~\ref{table:shapenet}. }
    \label{table:shapenet_emd_v}
\end{table*}

\begin{table}[!t]
    \begin{center}
        \small
        \begin{tabular}{ccc@{\hspace{0.2cm}}ccccc}
            \toprule
            VPL & CC & TP & IoU & CD${}_{s}$ & CD${}_{v}$ & EMD${}_{s}$ & EMD${}_{v}$ \\
            \hline
                       &            &            & $.403$ & $6.74$ & $5.59$ & $\mathbf{21.8}$ & $1.45$ \\
            \checkmark &            &            & $.490$ & $5.81$ & $3.93$ & $23.5$ & $0.76$ \\
            \checkmark & \checkmark &            & $.505$ & $4.81$ & $3.11$ & $\mathbf{21.8}$ & $0.84$ \\
                       &            & \checkmark & $.434$ & $5.29$ & $4.22$ & $23.3$ & $1.01$ \\
            \checkmark &            & \checkmark & $.508$ & $3.83$ & $2.75$ & $24.9$ & $0.72$ \\
            \checkmark & \checkmark & \checkmark & $\mathbf{.513}$ & $\mathbf{3.78}$ & $\mathbf{2.65}$ & $24.3$ & $\mathbf{0.70}$ \\
            \bottomrule
        \end{tabular}
    \end{center}
    \caption{Comparison of IoU, CD and EMD. This table is summary of Table~\ref{table:shapenet},~\ref{table:shapenet_cd_s},~\ref{table:shapenet_cd_v},~\ref{table:shapenet_emd_s}, and~\ref{table:shapenet_emd_v}. Subscripts $s$ and $v$ mean that points are uniformly sampled from surfaces and volumes respectively. VPL: proposed view prior learning. CC: class conditioning in the discriminator. TP: texture prediction. CD and EMD is lower is better. }
    \label{table:shapenet_summary}
\end{table}

In addition to intersection over union (IoU), Chamfer distance (CD) and earth mover's distance (EMD) are also often used for evaluation of 3D reconstruction. Table~\ref{table:shapenet_cd_s},~\ref{table:shapenet_cd_v},~\ref{table:shapenet_emd_s},~\ref{table:shapenet_emd_v}, and~\ref{table:shapenet_summary} show CD and EMD in the experiment of Table~\ref{table:shapenet} in the paper. We computed CD and EMD from points uniformly sampled from surfaces and volumes.

CDs and EMD${}_{v}$ correlate well to IoUs, which also validates our proposed method. EMD${}_{s}$ seems strange because EMD is greatly affected by spatial density of points and our method often generates spatially imbalanced surfaces. However, this imbalance hardly affects visual quality because it is often made by folding surfaces inside shapes. EMD${}_{v}$, which is computed from spatially uniform points, shows similar performance as other metrics.

%%%%%%%%%%%%%%%%%%%%%%%%%%%%%%%%%%%%%%%%%%%%%%%%%%%%%%%%%%%%%%%%%%%%%%%%%%%%%%%%
\subsection{Discriminators and optimization}

\begin{table}[!t]
    \small
    \begin{center}
        \begin{tabular}{cccc}
            \toprule
            Discriminator & Optimization & Texture & IoU \\
            \hline
            None                                 & -                 & & $.403$ \\
            Table~\ref{table:discriminators} (c) & Gradient reversal & & $.505$ \\
            Table~\ref{table:discriminators} (c) & Iterative         & & $.514$ \\
            Table~\ref{table:discriminators} (d) & Gradient reversal & & $\mathit{.403}^*$ \\
            Table~\ref{table:discriminators} (d) & Iterative         & & $\mathit{.403}^*$ \\
            \hline
            None                                 & -                 & \checkmark & $.434$ \\
            Table~\ref{table:discriminators} (c) & Gradient reversal & \checkmark & $.513$ \\
            Table~\ref{table:discriminators} (c) & Iterative         & \checkmark & $.510$ \\
            Table~\ref{table:discriminators} (d) & Gradient reversal & \checkmark & $\mathit{.434}^*$ \\
            Table~\ref{table:discriminators} (d) & Iterative         & \checkmark & $\mathit{.434}^*$ \\
            \bottomrule
        \end{tabular}
    \end{center}
    \caption{Evaluation of the discriminators in Table~\ref{table:discriminators} (c--d). ${}^*$No meaningful improvement was observed by tuning $\lambda_d$.}
    \label{table:appendix_experiment_discriminator}
\end{table}

Table~\ref{table:appendix_experiment_discriminator} shows the performances of the discriminators in Table~\ref{table:discriminators} (c--d) in single-view training. The discriminator in Table~\ref{table:discriminators} (d) does not work well in all cases. This table corresponds to the description in Section~\ref{sec:exp_dis_opt}.

%%%%%%%%%%%%%%%%%%%%%%%%%%%%%%%%%%%%%%%%%%%%%%%%%%%%%%%%%%%%%%%%%%%%%%%%%%%%%%%%
\subsection{Internal pressure in multi-view training}

\begin{table}[!t]
    \small
    \begin{center}
        \begin{tabular}{ccccc}
            \toprule
            Supervision & $N_v$  & Internal pressure & IoU \\
            \hline
            Single-view & $1$   &                   & $.387$ \\
            Single-view & $1$   & \checkmark        & $.403$ \\
            Multi-view  & $20$  &                   & $.648$ \\
            Multi-view  & $20$  & \checkmark        & $.652$ \\
            \bottomrule
        \end{tabular}
    \end{center}
    \caption{Effect of internal pressure loss.}
    \label{table:apppendix_internal_pressure}
\end{table}

In Section~\ref{sec:exp_priors}, we validated the effect of the internal pressure loss in single-view training. Table~\ref{table:apppendix_internal_pressure} shows that this loss is also effective in multi-view training. This experiment was conducted without texture prediction and view prior learning.

%%%%%%%%%%%%%%%%%%%%%%%%%%%%%%%%%%%%%%%%%%%%%%%%%%%%%%%%%%%%%%%%%%%%%%%%%%%%%%%%
\subsection{Loss functions of silhouettes}

\begin{table}[!t]
    \small
    \begin{center}
        \begin{tabular}{lc}
            \toprule
            Loss function & IoU \\
            \hline
            Multi-scale cosine distance (Eq.~\ref{eq:silhouette1}, $N_s = 5$) & $.575$ \\
            Multi-scale cosine distance (Eq.~\ref{eq:silhouette1}, $N_s = 1$) & $.567$ \\
            Interesection over union (Eq.~\ref{eq:silhouette2}) & $.552$ \\
            Sum of squared error & $\mathbf{.579}$ \\
            \bottomrule
        \end{tabular}
    \end{center}
    \caption{Comparison of silhouette loss functions.}
    \label{table:apppendix_silhouettes}
\end{table}

To compare two silhouette images, we used multi-scale cosine distance in Eq.~\ref{eq:silhouette1} and intersection over union (IoU) of silhouettes in Eq.~\ref{eq:silhouette2}. Table~\ref{table:apppendix_silhouettes} shows comparison of these loss functions in multi-view training ($N_v = 2$) on ShapeNet dataset. Additionally, sum of squared error of two silhouette images is compared. This result indicates that non-standard loss functions described in this paper are not so effective.

%%%%%%%%%%%%%%%%%%%%%%%%%%%%%%%%%%%%%%%%%%%%%%%%%%%%%%%%%%%%%%%%%%%%%%%%%%%%%%%%
\subsection{Modification of neural mesh renderer}

\begin{figure*}[!t]
    \begin{center}
    \includegraphics[width=387px,bb=0 0 527 183,trim={0.0cm 0.3cm 0.0cm 0.0cm},clip]{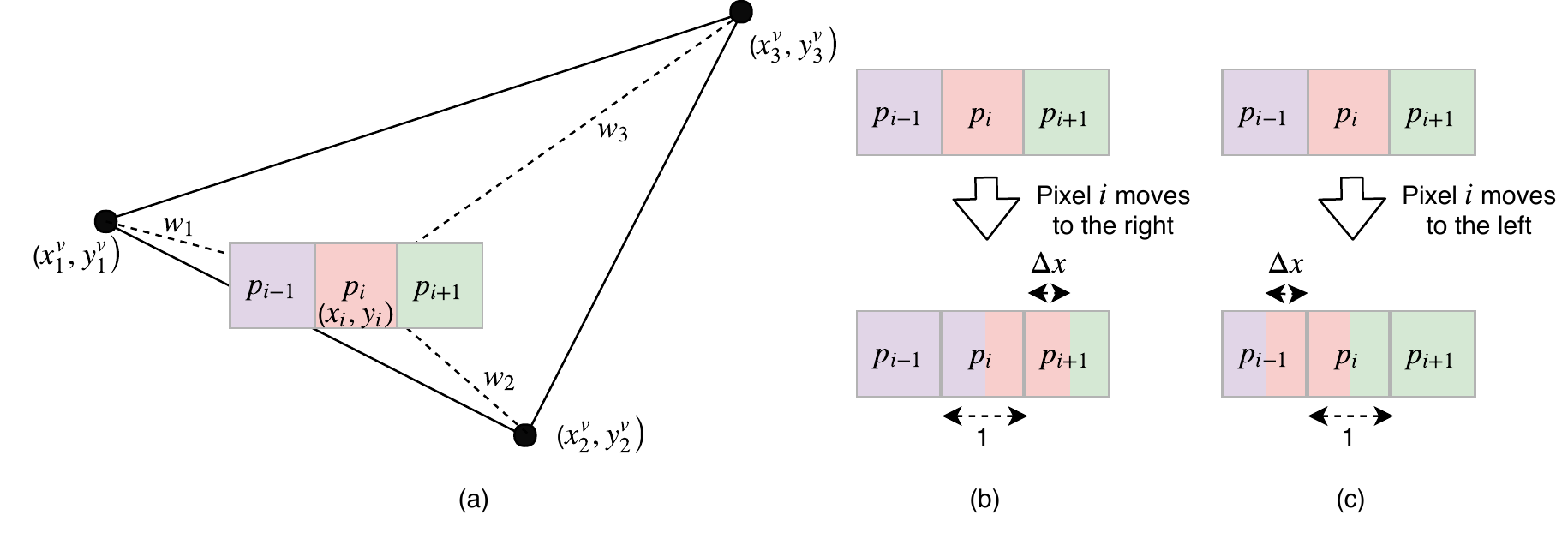}
    \end{center}
    \caption{Our assumptions on the differentiation of a renderer.}
    \label{fig:appendix_renderer}
\end{figure*}

In our implementation, we compute the differentiation of a renderer in a different way from~\cite{kato2018neural}. Their method is not stable when $\delta_i^x$ in~\cite{kato2018neural} is very small. Furthermore, the computation time is significant because a very large number of pixels is involved in computing the gradient with respect to one pixel. The approximate differentiation described in this section solves both problems.

Suppose three pixels are aligned horizontally, as shown in Figure~\ref{fig:appendix_renderer} (a). Their coordinates are $(x_{i-1}, y_{i-1})$, $(x_i, y_i)$, and $(x_{i+1}, y_{i+1})$, and their colors are $p_{i-1}$, $p_i$, and $p_{i+1}$, respectively. Pixel $i$ is located on a polygon, and its three vertices projected onto a 2D plane are $(x^v_1, y^v_1)$, $(x^v_2, y^v_2)$, and $(x^v_3, y^v_3)$. Then $(x_i, y_i)$ can be represented as their weighted sum $(x_i, y_i) = w_1 (x^v_1, y^v_1) + w_2 (x^v_2, y^v_2) + w_3 (x^v_3, y^v_3)$. Let $\mathcal{L}$ be the loss function of the network. When the gradient with respect to pixel $(\frac{\partial \mathcal{L}}{\partial x_i}, \frac{\partial \mathcal{L}}{\partial y_i})$ is obtained, the gradient with respect to the vertices of the polygon $(\frac{\partial \mathcal{L}}{\partial x^v_1}, \frac{\partial \mathcal{L}}{\partial y^v_1})$, $(\frac{\partial \mathcal{L}}{\partial x^v_2}, \frac{\partial \mathcal{L}}{\partial y^v_2})$, $(\frac{\partial \mathcal{L}}{\partial x^v_3}, \frac{\partial \mathcal{L}}{\partial y^v_3})$ can be computed using $w_1$, $w_2$, $w_3$, and the chain rule.

We assume that when pixel $i$ moves to the right by $\Delta x_i$, the pixel colors change, as shown in Figure~\ref{fig:appendix_renderer} (b). Concretely, the color of pixel $i$ changes to $p_i + (p_{i-1} - p_i) \Delta x$ and the color of pixel $i+1$ changes to $p_{i+1} + (p_{i} - p_{i+1}) \Delta x$. Then, $\frac{\partial p_i}{\partial x_i} = p_{i-1} - p_i$ and  $\frac{\partial p_{i+1}}{\partial x_i} = p_{i} - p_{i+1}$. Let $g_i$ be the gradient of the loss function back-propagated to pixel $i$. Concretely, $g_i = \frac{\partial \mathcal{L}}{\partial p_i}$. Then, the gradient of $x_i$ is
\begin{align}
    \frac{\partial \mathcal{L}}{\partial x_i}
    &= \frac{\partial \mathcal{L}}{\partial p_i}\frac{\partial p_i}{\partial x_i} + \frac{\partial \mathcal{L}}{\partial p_{i + 1}}\frac{\partial p_{i + 1}}{\partial x_i} \nonumber \\
    &= g_i (p_{i-1} - p_i) + g_{i + 1} (p_i - p_{i+1}) \nonumber \\
    &= (g^p_i)^{\text{right}}.
\end{align}

In the case where pixel $i$ moves to the left, we can compute the gradient in a similar manner. Thus,
\begin{align}
    \frac{\partial \mathcal{L}}{\partial x_i}
    &= \frac{\partial \mathcal{L}}{\partial p_i}\frac{\partial p_i}{\partial x_i} + \frac{\partial \mathcal{L}}{\partial p_{i-1}}\frac{\partial p_{i-1}}{\partial x_i} \nonumber \\
    &= g_i (p_i - p_{i+1}) + g_{i-1} (p_{i-1} - p_i) \nonumber \\
    &= (g^p_i)^{\text{left}}.
\end{align}

The problem is whether to use $(g^p_i)^{\text{right}}$ or $(g^p_i)^{\text{left}}$. When $x_i$ moves to the right, the decrease in $\mathcal{L}$ is proportional to $(d)^\text{right} = -(g^p_i)^{\text{right}}$. When $x_i$ moves to the left, the decrease in $\mathcal{L}$ is proportional to $(d)^\text{left} = (g^p_i)^{\text{left}}$. We define the gradient differently according to the following three cases.
\begin{itemize}
    \setlength\itemsep{0em}
    \item When $\max((d)^\text{right}, (d)^\text{left}) < 0$, the loss increases by moving the pixel $i$. Therefore, in this case, we define $\frac{\partial \mathcal{L}}{\partial x_i} = 0$.
    \item When $0 \leq \max((d)^\text{right}, (d)^\text{left})$ and $(d)^\text{left} < (d)^\text{right}$, the loss decreases more by moving pixel $i$ to the right. In this case, we define $\frac{\partial \mathcal{L}}{\partial x_i} = (g^p_i)^{\text{right}}$.
    \item When $0 \leq \max((d)^\text{right}, (d)^\text{left})$ and $(d)^\text{right} < (d)^\text{left}$, it is better to move pixel $i$ to the left. In this case, we define $\frac{\partial \mathcal{L}}{\partial x_i} = (g^p_i)^{\text{left}}$.
\end{itemize}

The gradient with respect to $y_i$ is defined in a similar way.

%%%%%%%%%%%%%%%%%%%%%%%%%%%%%%%%%%%%%%%%%%%%%%%%%%%%%%%%%%%%%%%%%%%%%%%%%%%%%%%%
\subsection{Experimental settings}

%%%%%%%%%%%%%%%%%%%%%%%%%%%%%%%%%%%%%%%%%%%%%%%%%%%%%%%%%%%%%%%%%%%%%%%%%%%%%%%%
\subsubsection{Optimizer}
We used the Adam optimizer~\cite{kingma2014adam} in all experiments. In our ShapeNet experiments, the Adam parameters were set to $\alpha=4e-4, \beta_1=0.5, \beta_2=0.999$. In the PASCAL experiments, the parameters were set to $\alpha=2e-5, \beta_1=0.5, \beta_2=0.999$. The batch size is set to $64$ in our ShapeNet experiments, and set to $16$ in our PASCAL experiments.

%%%%%%%%%%%%%%%%%%%%%%%%%%%%%%%%%%%%%%%%%%%%%%%%%%%%%%%%%%%%%%%%%%%%%%%%%%%%%%%%
\subsubsection{Encoder, decoder and discriminator}

\begin{figure}[!t]
    \begin{center}
    \includegraphics[width=181px,bb=0 0 246 297]{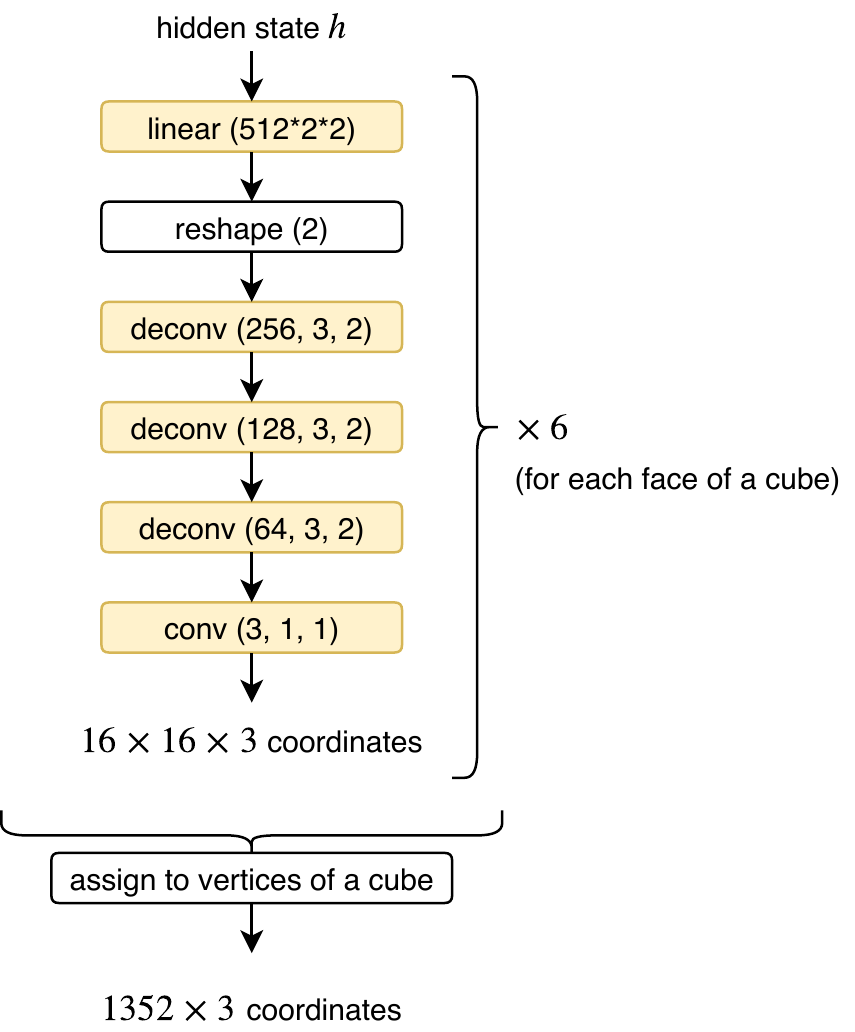}
    \end{center}
    \caption{Architecture of the shape decoder used in the ShapeNet experiments. The $16 \times 16$ vertices on each face of the cube are separately generated, and they are merged into $1352$ vertices. The dimension of the input vector is $512$. All linear and deconvolution layers except the last one are followed by ReLU nonlinearity.}
    \label{fig:appendix_shape_decoder_shapenet}
\end{figure}

\begin{figure}[!t]
    \begin{center}
    \includegraphics[width=75px,bb=0 0 102 138]{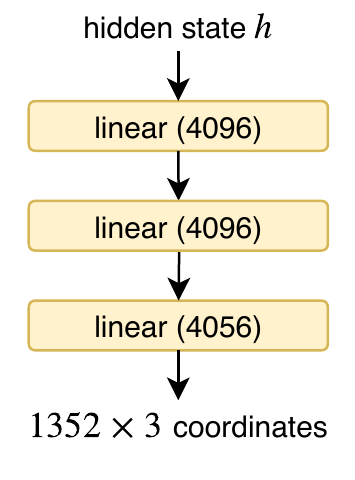}
    \end{center}
    \caption{Architecture of the shape decoder used in the PASCAL experiments. The dimension of the input vector is $512$. All linear layers except the last one are followed by ReLU nonlinearity.}
    \label{fig:appendix_shape_decoder_pascal}
\end{figure}

\begin{figure}[!t]
    \begin{center}
    \includegraphics[width=159px,bb=0 0 216 231]{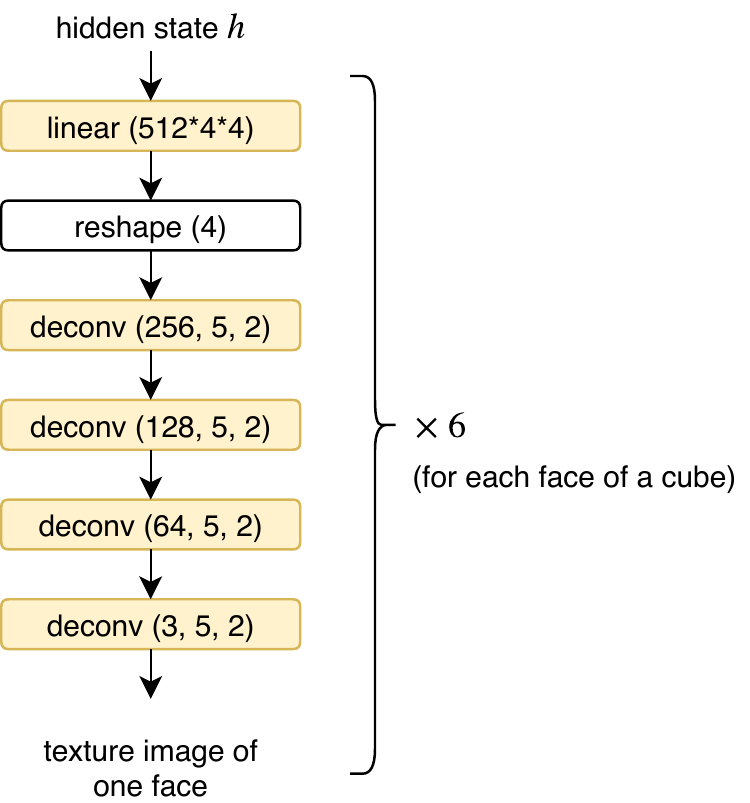}
    \end{center}
    \caption{Architecture of the texture decoder used in all experiments. A texture image of size $64 \times 64$ is generated separately for each face of a cube. The input vector has $512$ dimensions. All linear and deconvolution layers except the last one are followed by Batch Normalization~\cite{ioffe2015batch} and ReLU nonlinearity.}
    \label{fig:appendix_texture_decoder}
\end{figure}

\begin{figure}[!t]
    \begin{center}
    \includegraphics[width=117px,bb=0 0 159 243]{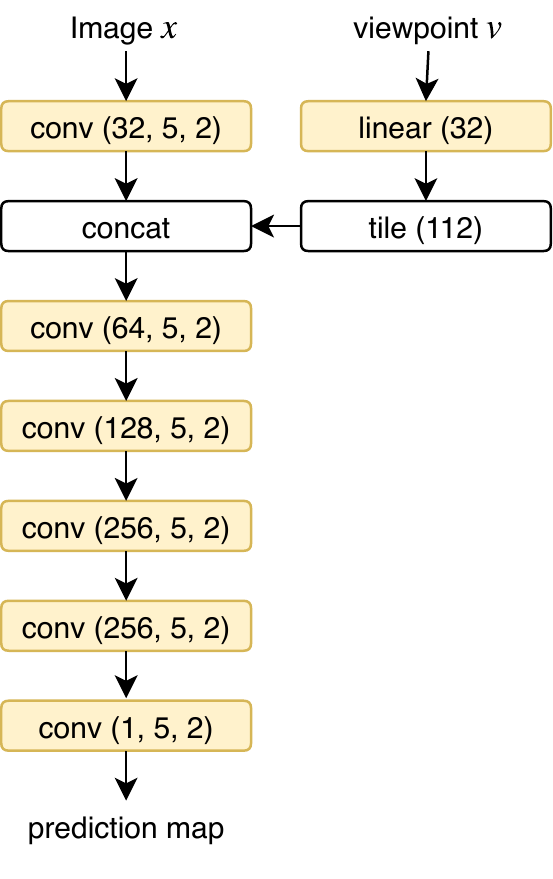}
    \end{center}
    \caption{The architecture of the discriminator used in the ShapeNet experiments. The size of the input image is $224 \times 224$. A viewpoint is represented by a three-dimensional vector of the elevation, azimuth, and distance to the object. Spectral Normalization~\cite{miyato2018spectral} is applied to all convolution and linear layers. All convolution layers except the last one are followed by LeakyReLU nonlinearity.}
    \label{fig:appendix_discriminator_shapenet}
\end{figure}

\begin{figure}[!t]
    \begin{center}
    \includegraphics[width=117px,bb=0 0 159 215]{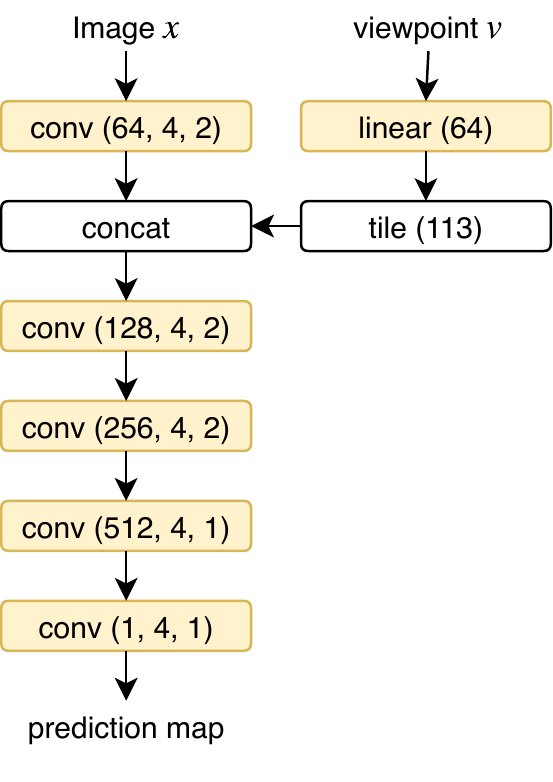}
    \end{center}
    \caption{Architecture of the discriminator used in the PASCAL experiments. The size of the input image is $224 \times 224$. A viewpoint is represented by a $3 \times 3$ rotation matrix. All convolution layers except the last one are followed by LeakyReLU nonlinearity.}
    \label{fig:appendix_discriminator_pascal}
\end{figure}

We used the ResNet-18 architecture~\cite{he2016deep} for the encoders in all experiments. The weights of the encoder were randomly initialized in the ShapeNet experiments. The weights were initialized using the weights of the pre-trained model from~\cite{he2016deep} in the PASCAL experiments.

We generated a 3D shape and texture image by deforming a pre-defined cube. The number of vertices on each face of the cube is $16 \times 16$, and the vertices on the edge of the cube are shared within two faces. The total number of vertices is $1352$. The size of a texture image on each face is $64 \times 64$ pixels. The shape decoder outputs the coordinates of the vertices of this cube, and the texture decoder outputs six texture images. Figures~\ref{fig:appendix_shape_decoder_shapenet} and ~\ref{fig:appendix_shape_decoder_pascal} show the architecture of the shape decoders used in the ShapeNet and PASCAL experiments. Figure~\ref{fig:appendix_texture_decoder} shows the architecture of the texture decoder used in all experiments.

Figures~\ref{fig:appendix_discriminator_shapenet} and ~\ref{fig:appendix_discriminator_pascal} show the architectures of the discriminators in the ShapeNet and PASCAL experiments.

The layers used in the architecture figures are as follows:
\begin{itemize}
    \setlength\itemsep{0em}
    \item $\text{linear} (a)$ is an affine transformation layer. $a$ is the number of feature maps.
    \item $\text{conv} (a, b, c)$ is a 2D convolution layer. The number of feature maps is $a$, the kernel size is $b \times b$, and the stride size is $c \times c$.
    \item $\text{deconv} (a, b, c)$ is a 2D deconvolution layer. The number of feature maps is $a$, the kernel size is $b \times b$, and the stride size is $c \times c$.
    \item $\text{reshape} (a)$ reshapes a vector into feature maps of size $a \times a$.
    \item $\text{tile} (a)$ tiles a vector into feature maps of size $a \times a$.
    \item $\text{concat} (\cdot)$ stacks two feature maps.
\end{itemize}

%%%%%%%%%%%%%%%%%%%%%%%%%%%%%%%%%%%%%%%%%%%%%%%%%%%%%%%%%%%%%%%%%%%%%%%%%%%%%%%%
\subsubsection{Other hyperparameters}

\begin{table*}[!t]
    \begin{center}
        \small
        \begin{tabular}{cccccccc}
            \toprule
            Training type & $N_v$ & Texture prediction & View prior learning & \#training iteration & $\lambda_c$ & $\lambda_d$  & $\lambda_p$ \\
            \hline
            single-view   & $1$           &            &            & $50000 $                  & -     & -      & 0.0001 \\
            single-view   & $1$           & \checkmark &            & $50000 $                  & $0.5$ & -      & 0.0001 \\
            single-view   & $1$           &            & \checkmark & $100000 $                 & -     & $0.2$  & 0.0001 \\
            single-view   & $1$           & \checkmark & \checkmark & $100000 $                 & $0.5$ & $2$    & 0.0001 \\
            \midrule
            multi-view    & $2,3,5,10,20$ &            &            & $25000 N_v$               & -     & -      & 0.0001 \\
            multi-view    & $2,20$        & \checkmark &            & $25000 N_v$               & $0.1$ & -      & 0.0001 \\
            multi-view    & $2,3,5,10,20$ &            & \checkmark & $50000 N_v$               & -     & $0.03$ & 0.0001 \\
            multi-view    & $2,20$        & \checkmark & \checkmark & $50000 N_v$               & $0.1$ & $0.3$  & 0.0001 \\
            \bottomrule
        \end{tabular}
    \end{center}
    \caption{Hyperparameters used in the ShapeNet experiments.}
    \label{table:hyperparameters_shapenet}
\end{table*}

\begin{table*}[!t]
    \begin{center}
        \small
        \begin{tabular}{ccccccc}
            \toprule
            Training type & Texture prediction & View prior learning & \#training iteration & $\lambda_c$ & $\lambda_d$  & $\lambda_p$ \\
            \hline
            category-agnostic &            &            & $15000$   & -      & -     & 0.00003 \\
            category-agnostic & \checkmark &            & $15000$   & $0.01$ & -     & 0.00003 \\
            category-agnostic &            & \checkmark & $50000$   & -      & $2$   & 0.00003 \\
            category-agnostic & \checkmark & \checkmark & $250000$  & $0.01$ & $0.5$ & 0.00003 \\
            \midrule
            category-specific &            &            & $5000$    & -      & -     & 0.00003 \\
            category-specific & \checkmark &            & $5000$    & $0.01$ & -     & 0.00003 \\
            category-specific &            & \checkmark & $40000$   & -      & $2$   & 0.00003 \\
            category-specific & \checkmark & \checkmark & $80000$   & $0.01$ & $0.5$ & 0.00003 \\
            \bottomrule
        \end{tabular}
    \end{center}
    \caption{Hyperparameters used in the PASCAL experiments.}
    \label{table:hyperparameters_pascal}
\end{table*}

Table~\ref{table:hyperparameters_shapenet} and Table~\ref{table:hyperparameters_pascal} show the number of training iteration and the weights of loss terms in ShapeNet and PASCAL experiments.

\end{document}